\documentclass[arts,article,submit,moreauthors,pdftex,10pt,a4paper]{mdpi2} 
\firstpage{1} 
\articlenumber{x}
\doinum{10.3390/------}
\pubvolume{xx}
\pubyear{2017}
\copyrightyear{2017}
\externaleditor{Academic Editor: Frederic Fol Leymarie}
\history{Received: date; Accepted: date; Published: date}

\Title{Choreographic and Somatic Approaches for the Development of Expressive Robotic Systems}

\Author{Amy LaViers $^{1}$\orcidA{}*, Catie Cuan$^{3,\dagger}$, Madison Heimerdinger$^{1,\dagger}$, Umer Huzaifa$^{1,\dagger}$, Catherine Maguire$^{3,4,\dagger}$, Reika McNish$^{5,\dagger}$, Alexandra Nilles$^{6,\dagger}$, Ishaan Pakrasi$^{1,\dagger}$, Karen Bradley$^{3,4,\ddagger}$, Kim Brooks Mata$^{7,8,\ddagger}$, Novoneel Chakraborty$^{2,\ddagger}$, Ilya Vidrin $^{9,10,\ddagger}$, and Alexander Zurawski$^{1}$} 

\AuthorNames{Amy LaViers, Catie Cuan, Madison Heimerdinger, Umer Huzaifa, Catherine Maguire, Reika McNish, Alexandra Nilles, Ishaan Pakrasi, Karen Bradley, Kim Brooks Mata, Novoneel Chakraborty, Ilya Vidrin, and Alexander Zurawski}  

\address{%
$^{1}$ \quad Mechanical Science and Engineering Department, University of Illinois at Urbana-Champaign\\
$^{2}$ \quad Aerospace Engineering Department, University of Illinois at Urbana-Champaign\\
$^{3}$ \quad Independent dance artist or professional\\
$^{4}$ \quad Laban/Bartenieff Institute of Movement Studies\\
$^{5}$ \quad Kinesiology and Community Health Department, University of Illinois at Urbana-Champaign\\
$^{6}$ \quad Computer Science Department, University of Illinois at Urbana-Champaign\\
$^{7}$ \quad Dance Program, University of Virginia\\
$^{8}$ \quad Integrated Movement Studies\\
$^{9}$ \quad Centre for Dance Research, Coventry University\\
$^{10}$ \quad Department of Theatre, Dance, and Media, Harvard University}

\corres{Correspondence: alaviers@illinois.edu; +1-217-300-1486}

\firstnote{These authors contributed equally to this work.} 
\secondnote{These authors contributed equally to this work.}

\abstract{ 
As robotic systems are moved out of factory work cells into human-facing environments questions of choreography become central to their design, placement, and application.  With a human viewer or counterpart present, a system will automatically be interpreted within context, style of movement, and form factor by human beings as animate elements of their environment.  The interpretation by this human counterpart is critical to the success of the system’s integration: ``knobs’’ on the system need to make sense to a human counterpart; an artificial agent should have a way of notifying a human counterpart of a change in system state, possibly through motion profiles; and the motion of a human counterpart may have important contextual clues for task completion.  Thus, professional choreographers, dance practitioners, and movement analysts are critical to research in robotics.  They have design methods for movement that align with human audience perception; they can help identify simplified features of movement that will effectively accomplish human-robot interaction goals; and they have detailed knowledge of the capacity of human movement.  This article provides approaches employed by one research lab, specific impacts on technical and artistic projects within, and principles that may guide future such work.  The background section reports on choreography, somatic perspectives, improvisation, the Laban/Bartenieff Movement System, and robotics.  From this context methods including embodied exercises, writing prompts, and community building activities have been developed to facilitate interdisciplinary research.  The results of this work is presented as an overview of a smattering of projects in areas like high-level motion planning, software development for rapid prototyping of movement, artistic output, and user studies that help understand how people interpret movement.  Finally, guiding principles for other groups to adopt are posited.
}


\usepackage{wrapfig}
\usepackage{subcaption}

\begin{document}


\section{Introduction}\label{intro}

\begin{quote}
\textit{\textbf{Domin}: So young Rossum said to himself: Man is a being that does things such as feeling happiness, plays the violin, likes to go for a walk, and all sorts of other things which are simply not needed.\\
\textbf{Domin}: No, wait.  Which are simply not needed for activities such as weaving or calculating.  A petrol engine doesn't have any ornaments or tassels on it, and making an artificial worker is just like making a petrol engine.  The simpler you make production the better you make the product.  What sort of worker do you think is the best?\\
\textbf{Helena}: The best sort of worker?  I suppose one who is honest and dedicated.\\
\textbf{Domin}: No.  The best sort of worker is the cheapest worker.  The one that has the least needs.  What young Rossum invented was a worker with the least needs possible.  He had to make him simpler.  He threw out everything that wasn't of direct use in his work, that's to say, he threw out the man and put in the robot.  Miss Glory, robots are not people.  They are mechanically much better than we are, they have an amazing ability to understand things, but they don't have a soul.  Young Rossum created something much more sophisticated than Nature ever did -- technically at least!}
\end{quote}
\vspace{-.2in}
\begin{flushright}
-- RUR (Rossums’ Universal Robots) \cite{vcapek2004rur}
\end{flushright}

Bringing together choreography and engineering raises a question about whether both fields are concerned with the same academic values and inquiries. That is, are they motivated by the same organizing principles and can their practices be complementary, and further mutually beneficial, when brought together? 
In the above excerpt from the play where the term ``robot'' was coined, the immediate intertwining of robotics and the arts can be observed.  From the first moment the concept of a ``robot'' was conceived there was an idea injected that movement could be divided between its efficient, functional parts and its extraneous, expressive parts.  Thus, the idea, of efficiency, and function, has permeated robotics since the word was introduced in a 1920 play \cite{vcapek2004rur}.  Long before they were called robots, automatic mechanical machines, automata, have existed since antiquity and as Truitt writes ``are mimetic objects that dramatize the structure of the cosmos and humankind's role in it'' \cite{truitt2015medieval}, pointing to the long held mysticism and romance associated with complex machines.

At the same time, a choreographer Rudolf Laban (1879-1958) was working to establish a system for movement, now called the Laban/Bartenieff Movement System (LBMS), that helped codify learning from work in the arts and helped frame a rejection of industrialization and return to nature  \cite{bradley2008rudolf} .  Thus, both the \textit{idea} of a robot and this somatic body-based movement system that will be leveraged heavily in the methods presented here are in part a reaction to the industrial revolution where mechanical efficiency -- and its abstract ideal -- was taken to hyperbolic levels not previously seen.  
Dancers’ technique and training often reflects the ethos of developing efficiency through iteration of motor tasks, continually refined and streamlined. But, dance performance requires the use of something ``more'' -- maybe the same thing Domin and Helena discuss as missing from the concept of a robot.    

Thus, as much as physical phenomenon, the distinction between human and robot movement is a philosophical distinction.
In broad, reductive strokes, we argue that both roboticists and choreographers aim to do the same thing:  to understand and convey subtle choices in movement within a given context. One of the fundamental emergent questions from this categorization is about the lens through which constituents in each field methodologically approach how to generate, interpret, and reproduce different movement from the idiosyncratic to the robotic.  
Consider two branches of inquiry, one concerned with internal experience and the other with external measurement as a model for understanding the differences in assumptions generated from each field. 

Different branches of philosophical inquiry provide critical, albeit artificial, lenses by which to address this complex question. Two such branches, rhetoric and phenomenology, are useful in understanding subjective and objective experience, though notably from different angles. Rhetoric, at least in the classical (Aristotelian) sense, investigates ``the available means of persuasion'' in any given situation \cite{kennedy2006rhetoric}. Outside of ancient Greek life, contemporary rhetoric can be understood as a practice to examine the function, efficiency, and efficacy of persuasive communication—starting with linguistic, and moving to the visual, bodily, energetic, posthuman, and so on \cite{barnett2016rhetoric}. Alternatively, phenomenology provides a lens through which to understand subjective, first-person experience. Championed by Edmund Husserl at the turn of the twentieth century, phenomenology urges a ``bracketing'' of experience such that an individual approaches stimuli as if for the first time \cite{husserl2012ideas}. Phenomenology and rhetoric are not, strictly speaking, incongruous. One can approach the creation of rhetorical artifacts (i.e. speeches, visualizations, choreographies, etc.) phenomenologically, just as one can probe first-person experience rhetorically.  These lenses are artificial constructs -- they offer a way by which to view the world but their limitations are neither absolute nor universal.  \cite{hawhee2004bodily}  

Returning to the questions of engineers and choreographers, a major problem can occur when either one becomes too beholden to their value systems. For engineers, the problem occurs when valuing efficiency and function (i.e. rhetoric) trump user experience (i.e. phenomenology). Conversely, for choreographers, the problem occurs when first-person experience (of the performers) is not observable by an external viewer (and therefore, maybe one day, quantified). As with any practice, balance is imperative. To be able to address complex questions pertaining to generating, interpreting, and reproducing movement, relying too heavily on rhetorical or phenomenological aims can be limiting in a way that be invisible or simply not salient for the agents. This is one reason, among many, to support bringing engineers and choreographers together into one shared space, and, for a robotics lab, to employ phenomenologically motivated practice in our research. In bridging the first-person experience with questions of efficacy and efficiency, we can move closer to producing more \textit{expressive} robotic systems.

In the largest sense, the purpose of the methods presented here is to explore what human bodies can express in movement, what ideas machines can express in movement and whether they are the same things. Can a robot expressively communicate?  If we assert that all movement is expressive (which we do), then can a robot express the same set of things as a human?  The other big idea is to utilize the choreography of the body itself, and its expressive capabilities, to ``move''  engineers into an embodied experience that could inform their work both in design and context creation.  Meaning making through movement -- engineering through body.

Thus, our approach is pragmatic: we want to understand the phenomenon of how people create such vastly varied motion profiles that communicate complex intent.  This knowledge is contained inside body-based movement training and somatic practice\footnote{Somatic approaches, being distinguished from a broad category of body-based approaches, allow us to extract knowledge from experience the of body from an internal perspective, which is called “soma”, a distinct idea from the body itself \cite{eddy2009brief}.} where practitioners hone their own movement capabilities by expanding their array of choices.  There are a multitude of body-based practices \cite{foster2004somatic}, including very well known ones, such as Alexander Technique \cite{alexander1990alexander}, Feldenkrais Technique \cite{feldenkrais1972awareness}, Body Mind Centering \cite{cohen2012sensing}, Pilates Method \cite{latey2001pilates} and Contact Improvisation \cite{pallant2006contact} to name just a few, but they all share some commonalities in their approach to experiencing the body from an internal, rather than external, perspective \cite{hanna1980body}.  External methods, typically employed in the sciences and in engineering such as motion capture, photography, force plates, and the like, can work to document the result of a movement pattern but do not have access to choices made by a human in focus, motivation, sensation, memory, prior muscle patterning (and re-patterning), etc.  The practice of honing these choices is one of \textit{embodiment} -- a body of knowledge that cannot be \textit{known} but only \textit{moved}.

Indeed in philosophy and research in the arts a question can be posed of what kinds of ideas can be expressed through movement \cite{exempli} and what kind of learning is inherently kinesthetic \cite{abrahamson2004embodied,goldman2010increasing,lindgren2013emboldened}.  A field of work known as choreology \cite{hall1964dance,sutil2015motion} defines methodology that enables deeper thinking about what defines movement.
In robotics, the field has typically motivated, without much deep thought, a separation between functional, task-based movement and ``expressive motion'' profiles only useful for the purposes of human responses.
LBMS frames efficiency as the process of selecting from all that is possible, the clearest choices for the richness of human physical expression to occur. Recognize the word play that happens when one considers ``the efficiency of enjoying the task and being in relationship with another'' or ``the expressiveness of a clear, decisive, precise movement through space''. Thus, \textit{all} movement is ``expressive'' \textit{and} ``functional'' -- and the tools honed through time spent in a dance studio will help to create more expressive robotic \textit{systems} that have broader, more successful functionality in dynamic environments.

This approach has been adopted through several diverse activities organized through the Robotics, Automation, and Dance (RAD) Lab.  This lab was at the University of Virginia (UVA) from 2013-2015 and is now at the University of Illinois at Urbana-Champaign (UIUC) since 2015.  In this article, the author group writes as one body (although we are from multiple perspectives and experiences) for clarity.  A list of specific activities, along with the names of collaborators, is given below.

\begin{itemize}
\item Between 2013 and 2017 Catherine Maguire offered five workshops in LBMS for the RAD Lab.  These ranged from one-on-one sessions to group sessions with all lab members and sponsored from the lab start up packages at UVA and UIUC as well as a grant from the UVA Data Science Institute. 
\item In Fall 2014 Kim Brooks Mata and Amy LaViers co-developed and co-taught DAN 3559 / ENG 3501 Electronic Identity and Embodied Technology Atelier, cross-listed between the dance and engineering programs at the University of Virginia and funded by the Jefferson Trust.
\item As part of a research project, ``Choreography of Platform-Invariant Motion Primitives'' Amy LaViers organized two three day training workshops at UIUC.  The first, in June 2016, was co-developed and co-taught with Catherine Maguire and Karen Studd; the second, in June 2017, was co-developed and co-taught with Catherine Maguire, Catie Cuan, and Riley Watts.  In addition to travel expenses covered, the facilitators were paid at one of two hourly rates (one for consulting and a higher rate for teaching).  These workshops were sponsored by DARPA grant number D16AP00001.
\item Karen Bradley advised Amy LaViers' CMA thesis project entitled ``Preparing to Cook with Eric: Function and Expression Within Recipe'', which was completed in August 2016 through the Laban/Bartenieff Institute of Movement Studies (LIMS).
\item Amy LaViers organized a workshop at Robotics: Science and Systems (RSS) in 2016 entitled ``Let's Move: Embodied Experience and Movement Observation for Roboticists''.  The workshop featured Elizabeth Jochum, Heather Knight, and Kayhan Ozcimder as speakers.  Attendees included participants from industry and academia including CMU, JPL, U Mich, Duke, WPI, Google, Sphero, UIUC, Princeton, and Aalborg University. 
\item Amy LaViers co-organized an invited, follow on workshop in 2017 for RSS with Kayhan Ozcimder entitled ``Experimenting with Movement Observation''.  Attendees included participants from MIT, UIUC, Princeton, and U Mich.
\item In April 2017 Amy LaViers attended Art, Tech, Psyche III at Harvard University where she also spent extended time meeting with Ilya Vidrin and Riley Watts discussing the phenomena of partnering in dance; these conversations have led to an inprogress independent study project for a master's student in mechanical engineering at UIUC.
\item In June 2017 Catie Cuan, a New York based choreographer, spent a month in residence at the RAD Lab in Urbana, IL.  As part of this residency she taught classes to lab members and developed an artistic piece alongside Amy LaViers, Ishaan Pakrasi, and Novoneel Chakraborty.  The piece ``Time to Compile”, and was presented as a work in progress on June 30, 2017.  This work is sponsored by the lab start up package and NSF grant number 1528036.  User studies are in progress using this material to modulate perception of in home robots.
\item In addition to training and education for lab members, Catherine Maguire has consulted on two research projects where her expertise in movement theory and observation have been integral parts of the technical output in the lab.  In addition to co-authoring resulting papers, she has been paid an hourly consulting rate on these projects, funded by NSF grant numbers 1528036 and 1701295. 
\item In Spring 2017 a new course ME 598: High-level Robotic Control and Movement Representation was offered at UIUC.  The course emphasized embodied movement exploration, writing, and interdisciplinary research.  An ``Expressivity Expansion Pack'' for the Robot Design Game\footnote{http://radlab.mechse.illinois.edu/2017/06/14/robot-card-game/} was produced as a formal output of this class.
\item Over the years two regular lab activities have been developed that support this line of work in the group in addition to the group's weekly lab meeting.  We call them ``writing hour'' and ``movement hour''.  In each, we take time to practice descriptive writing and embodied movement exploration (led by Amy LaViers or a guest facilitator like Catie Cuan who helped establish ``movement hour'') in order to supplement the traditional education students receive in engineering.  Students in the group who have contributed as co-authors here are Novoneel Chakraborty, Madison Heimerdinger, Umer Huzaifa, Reika McNish, Alexandra Nilles, Ishaan Pakrasi, and Alexander Zurawski.
\item Lab outreach activities also benefit from these workshops.  The lab holds outreach activities for students in age ranges from elementary to high school.  All of these activities work to showcase robotics in an accessible manner and feature embodied movement exploration as well as quantitative \textit{and} qualitative objective description of movement of machines \textit{and} humans.
\end{itemize}

\subsection{Choreography as Body-Based Research}


Why movement courses for engineers?  
Collaboration with dance practioners brings largely universal movement principles into the coterie of source inspiration and awareness for engineers.  Movement knowledge empowers lab students to find context and meaning in seemingly commonplace movements.  How does a hand wave meaning ``hello'' differ from a hand wave of ``move out of the way''?  When tempo and frequency change, is the movement ``the same''?  Importantly, movement awareness shows how narrative is drawn from any scale of movement in space.  For instance, a Roomba floor robot skids along the floor in a continuous motion.  Possible assigned narratives include subservience, a snake-like creep, and a treadmill.  How does this type of movement and resulting narrative affect the perception of the robot overall?  The Roomba takes up space within an owner’s home and therefore becomes part of their personal narrative.  How will that owner move differently or rearrange their space as a result of the robot?  Undoubtedly, this new addition alters their daily story.  It is critical engineers are given human-centered movement tools to understand these alterations and perceptions that stem from their designs.

The methods utilized by choreographers help create meaningful movement\footnote{Note that meaningful movement may not necessarily be narrative.} and organize the execution of large-scale performances, which may not take on citations in academic journals, but are nonetheless important ways in which society organizes knowledge and makes sense of experience.  For example, Rainer's ``Trio A'' initially didn't look like a dance because she used pedestrian-style movements that do not have the typical virtuousic leg extensions, etc. associated with dance \cite{rainer1966quasi,lambert2008being}; this work was heavily influenced by the context of the larger social movements connected with her and others working at the Judson Memorial Church at the same time \cite{banes1983democracy,banes2011terpsichore}.  Thus, the innovative, groundbreaking movement profiles put forth by Rainer puzzled the dance community and audiences but revealed that pedestrian movement is already expressive and masterful \cite{exempli}.  Innovation in any discipline is often associated with disorientation of existing patterns; here it is specifically new movement patterns that help express a greater variety of ideas.  Moreover, the philosophy embodied by her work \cite{rainer2006no} informed minimalism in other art forms like fashion, music, and theater \cite{copeland1993dance,lambert1999moving}.

External technologies can augment human movement designers, providing tools such as randomness, or immediate visualization, which can inspire new creative directions as described in \cite{improvhand}.  One well-known adopter of such external technologies to aid in motion design was Merce Cunningham \cite{copeland2004merce}. For instance, he would use coin flips and the I Ching to randomly determine the sequence of movement in his choreography, and said of this technique, ``the feeling that I have when I compose in this way is that I
am in touch with a natural resource far greater than my own personal inventiveness could ever be" \cite{cunningham1997impermanent}. Similarly, Cunningham also adopted computer technology, such as the LifeForms
software \cite{calvert1993evolution,schiphorst1993case} which is a graphical tool for choreographers to compose dances using animated representations of human dances. 

While robotics and dance are both concerned with physicalized movement through space, there are fundamental differences between their process and execution that manifest valuable learnings. The process of dance making is often elusive and always unique to the individual.  One choreographer may begin with musical inspiration and create phrases that closely match the musical score.  These phrases may be constructed immediately after a first listen or much later in a new context.  One choreographer may create an entire piece of choreography without sound and add it only at the end of the process.  Another may work collaboratively with a composer from beginning to end.  Some choreographers may choose to work from a theme and engage their dancers in the creative process by asking that they participate in the creation of movement material alongside the choreographer.  Another could have a conceptual inspiration and ruminate over those themes for weeks before finally starting movement phrasing.  In all cases, the movement sequence could still appear in just a few minutes and be used in a performance.  

These examples show the overwhelming multitude of ways of working in the arts and how they can be undermined by engineers who work with tools that typically eliminate the possibility of iterating on designs so quickly.   Choreography is an iterative process requiring going back and reworking choices and incorporating feedback regarding the ``success'' of each of the choices in context and how they support an artists intent and how they elicit particular responses in audience members. This body-based research is thus not done after a three minute improvisation, though there are certainly serendipitous moments in rehearsal that can lead to major leaps forward in the creative process; indeed, sorting through these choices is as arduous as crafting a line of code. The research accomplished inside the choreographic process includes the engagement of other bodies and requires some form of collaborative investigation no matter what the degree of creative input the performers might have with the choreography they are still adding their artistry and skill in the manifestation and performance of the work. 

The nature of each process results in different amounts of time needed to iterate on a given design, which, in making a metaphor with computer architecture, we've come to call ``compile time''. The compile time for a choreographer could be considered the weeks of rumination, or the first listen, before there is some movement phrase to be shown.  If forced to create a short dance in three minutes, a choreographer could certainly have something finished.  Certain choreographers are known for presenting improvisation on stage where the compile time is arguably close to zero.  Engineering, in contrast, rarely ever results in a functioning ``performance'' or design loop on the first try.  Iterations and bug fixes are necessary, hours-long endeavors before finalization.  In the similar three minute example, it would be rare to see a working design loop.  This contrast in compile times is valuable as researchers can create and discuss movement creations immediately, through the media of their bodies rather than arduous scripts in heavy frameworks like ROS (the Robotic Operating System; discussed further in Section \ref{alli_sec}).

\subsection{Improvisational Technology}

Improvisation is an important process used to support choreography (movement design) and dance training and utilized in performance that is distinct enough to be called out on its own. It also employs a large role in the somatic approaches taken by the RAD Lab -- activities such as movement hour incorporate individual and group improvisation activities (described in Section
\ref{methods}). These activities are tools for improvising \textit{in the body}, each highlighting different aspects of movement and providing important opportunities to explore and learn about movement. Just as we can improvise with external technologies -- such as musical instruments or software programs -- we can improvise using the body. 
William Forsythe has worked extensively to explicate these strategies, which he employs in training, choreography, and performance, \cite{forsythe2012improvisation}.

One improvisational technique from dance is contact improvisation, which is taught at UIUC as DAN 103 Contact Improvisation and DAN 259/459 Contact Improvisation for Musicians, Actors, and Dancers. As the name suggests, this form of dance is improvised, sometimes to music, and emphasizes contact (with other humans, walls, floors, etc). This embodied approach to studying movement improvisation can provide valuable experience to roboticists interested in the spontaneous design of movement for bodies which interact with their environment.  Further, a mature improvisational practice develops deep understanding of what choices are available and which may be generated in response to an exigency or problem \cite{goehr2014improvising}.
As roboticists, improvisational techniques are useful at two levels. First, if a human is designing robot motion (such as a gait, or the path a robot should take through a warehouse), it is useful to be able to \textit{improvise} instructions to the robot - allowing the designer to quickly iterate on their instructions and get
feedback immediately about how their instructions are interpreted on the robot platform. On the other hand, we are moving toward robots that need to ``improvise" their own movement -- for example, autonomous vehicles that must react to the actions of human drivers around them. One exercise in contact improvisation is to walk in a group around a room, with each person improvising their path, and gradually increasing the speed until everyone is running around the room and must manage to avoid collisions by dodging other people at the last second, or making contact with them in a way that avoids injury. As an exercise for roboticists, this helps us understand and analyze techniques for communicating intent, and using momentum to maintain control and avoid damage -- techniques very relevant to problems studied by roboticists.

Another important concept from improvised dance is an embodied understanding of creative flow and an appreciation for the complexity of the human body. As most people have experienced, improvising dance is difficult. It does not feel natural at first. But with practice, it becomes easier to slip into a mental state where there is no self-consciousness and the dance is done as an instinctive reaction to music, previous movement, or the movements of other people around you.
Setting an intention to move beyond known or familiar movement patterns supplements this work \cite{peters2009philosophy}.
This embodied experience often leads to a deep appreciation of the complexity of the human body and movement: when improvising, we choose naturally from an extremely large set of possible movements. The calculations done by the human nervous system clearly outpace our best digital optimizers and controllers. 
A practice in improvisation also gives an appreciation for the usefulness of constraints: trying to improvise a completely unconstrained dance is much more difficult than improvising "walking forward as if you are moving through knee-high water," for example.
In robotics constraints are similarly useful: general-purpose robots are much more difficult to engineer than ones which only perform or respond in specific contexts.
Indeed, after careful consideration of the body the task of creating a robot which compares in expressivity begins to seem nearly impossible.
Analytical models, which acknowledge the nonlinearity\footnote{We could offer an entire discussion around the use of this word.  Inside systems and control, it is well-acknowledged that almost no system is linear.  The difficulty in working with nonlinear models necessitates linear models, which do well to predict many systems inside narrow operating regimes.  Yet, dance professionals tend to view this necessary simplification as a failing of the field -- to the point that the idea of ``linear'' vs. ``nonlinear'' thinking is used as a belittlement toward engineers in the same way that ``technical'' and ``nontechnical'' get used, in the same vacuous manner, by engineers to knock those specializing in qualitative methods like dance scholars.  In this article, we want to 1) point out how unprofessional and unproductive such baseless attacks, from both sides, are and 2) emphasize the usefulness of simplified quantification and the technical expertise involved in qualitative and embodied practices.  We've found that acknowledging and balancing these views is necessary for meaningful collaboration.} of something as complex as improvisation, have been used to try and describe this behavior inside structured improvisations in performance as in \cite{ozcimder2016investigating} or to generate diverse movement through chaotic models \cite{chaos}. Our goal in this article is to extract embodied understanding that can guide in the development of expressive tools, rather than debating the possibility of biological processes in machines.


\subsection{Laban/Bartenieff Movement Studies (LBMS) and Certified (Laban) Movement Analysts}

One tool set that can be used to describe (and similarly generate) novel movement styles like Rainer's is called the Laban/Bartenieff Movement System (LBMS) \cite{laban1971mastery,laban2,laban0,laban1966choreutics,bartenieff1980body,hackney1998making,studd2013}.  This work is mainly communicated through certification programs in its institutes, the Laban/Bartenieff Institute of Movement Studies (LIMS) and the Integrated Movement Studies (IMS), where successful participants become Certified Movement Analysts (CMAs) and Certified Laban/Bartenieff Movement Analysts (CLMAs), respectively.  The emphasis on in-person training is part of the programs' philosophy: the material cannot be ``understood'' without bodily, or embodied, participation.
A system of movement analysis taught in these programs and initiated by Laban and his student Irmgard Bartenieff, LBMS is utilized in many professional contexts, such as therapy, consulting, and research, in addition to dance and choreography. 
 Many faculty members in dance departments across the world hold certifications through these institutes; however, there are large differences between this work that make interdisciplinary collaborations challenging.  Unlike other academic disciplines, this field does not publish regularly \cite{groves2007talking} and does not offer doctoral degrees.  This is a challenge not only in creating tenured positions for members in this group but also for accurate transferal of knowledge.  It is hard to find current or complete documentations of this work in print.  
Nevertheless, this body-based approach to research has much to offer the field of robotics. 

The work in LBMS takes into account the larger patterns of human movement identified as Thematic Dualities. One such theme (of four major ones identified) is Function/Expression (F/E), which clarifies that the need for expressive robotic systems is a practical, functional pursuit. It is important to note that finding patterns is in fact the primary experience of our body moving in relationship to our environment. It is a process of differentiation that ultimately allows for synthesis -- understanding that while function and expression can be perceived as opposites, they are in fact inseparable.  In the LBMS curriculum, these themes are described with a mobius strip topology, indicating that the ideas are one and the same.  This is a primary influence on our point of view regarding expressive robotic systems.

Further, LBMS provides a series of interrelated, qualitative lenses, termed Body (answering ``What?'' about movement), Effort (``How?''), Space (``Where?''), and Shape (``Why?''), which contribute to notation systems, for practitioners to use in finding pattern and meaning in movement.  The overlaps in these categories are explicated through Affinities (part of Laban’s theory of Space Harmony \cite{laban1971mastery}).   
The idea of Affinities is simply that there are large patterns that can be identified through relationships among Body, Effort, Shape and Space. For example, certain kinds of expression tend to occur in particular directions in Space. This relationship is actually rooted in the body’s design, (once again foregrounding the idea of body as basis and movement as a form of knowing) for example Light Weight Effort is linked to the Center of Levity, located in the Upper Body -- (up) -- and Strong Weight Effort is linked to the Center of Gravity located in the Lower Body -- (down). As humans we associate particular movement expressivity with certain tasks and series of changes in spatial location that allows us to context meaningful communication.  Note, that this context, most generally, is comprised of the summative experience of a lived life -- thus, meaning making in movement is a function of culture, prior experience, as well as immediate situational context. 

\subsection{Prior Work in Expressive Robotics}

Motion for robots that is termed ``expressive'' has been used to describe movement used in social and human-facing settings.  From Breazeal's Kismet \cite{breazeal2004designing}, which employed imitations of facial features in order to create engaging interactions through a mechanical system, to Knight's candy delivering systems \cite{knight2015taking}, it has been shown that genuine engagement between human and robots can be created through design of hardware, motion profiles, and contextual interaction.  
This motion is often described with affective labels \cite{breazeal1999context,knight2012acting,knight2014expressive} or, similar to Domin in the play that spawned the term ``robot'', as motion features orthogonal to function \cite{knight2015layering}.  Extensive work in animation has explored similar veins of work \cite{chaos,brand2000style,liu2005learning,torresani2007learning,gillies2009learning}.  For example, \cite{etemad2016expert} validates motion profiles via lay viewers under labels of ``happy'', ``sad'', ``tired'', ``energetic'', ``feminine'', and ``masculine''.  LBMS has also been utilized in several such academic publications in the field of robotics \cite{patla1982aspects,hudak,rett2007human,rett2008laban,masuda2009emotion,lourens2010communicating,masuda2010motion,knight2014expressive,knight2015layering,barakova2015observation,knight2016laban}.  
Work in \cite{gielniak2010stylized} has shown that just adding \textit{variation} to movement makes it seem more human-like to human viewers.  Other work has focused on functional interpretation, termed \textit{legibility} and \textit{predictability}, in narrow contexts \cite{dragan2013legibility}.

The affective domain of modern psychology encompasses the experience of feeling or emotion.  Affective states can be decomposed into three dimensions: valence, motivational intensity, and arousal. Russell developed a model of affect that is used to map affective states to a two dimensional chart using ratings of valence and arousal \cite{russell1980circumplex}. This model has been used in an array of applications, including affective classification of blog posts \cite{paltoglou2013seeing} and images with affective ratings of valence and arousal \cite{dan2011geneva,kurdi2017introducing}.  This portends a stark contrast between the approach used by psycologists and roboticists in labeling stimuli.  Where psycologists are using a multi-dimensional model, where emotions like ``happy'' and ``sad'' may be placed inside a less loaded and personal parameter space (valence, arousal, and motivational intensity or dominance), roboticists use specific, emotive terms that may differ in distinct individuals and contexts.

Studies in the psychology of adoption of robotic technologies in the household outline the importance of social intelligence and perceived animacy. These studies show that by expressing character traits, a robotic system can imply levels of social intelligence that increases the likelihood of adoption. \cite{hendriks2011robot}, \cite{young2011evaluating} tell us that it is natural for humans to try and extract information from robotic actions, subsequently attributing intentionality to robot movement characteristics and decision making. In \cite{forlizzi2006service}, for instance, user study participants describe the movements of a robot vacuum cleaning system, the Roomba as ``cute'' or ``pathetic'', even though such a correlation may not have been intentioned. There are also instances where people name their Roomba robot, thus giving it an added social identity. The achievement of social assimilation albeit by coincidence and not by intention, prompts the user to associate decisions made by the robot to its personality traits, as opposed to the functional algorithm that determines its movements. In \cite{darling2015empathic} researchers explore the relationship between emphatic concern and the effect of stories in the interaction with robotic systems. This is done through a user study, where participants are asked to strike and destroy a robot insect, with a mallet. Some of the miniature robot insects are given a backstory as described to participants, and some are not. Results of the study show that people are less likely to strike the insects with a backstory, thus proposing a relationship between empathy towards robots and the existence of a priming backstory. This indicates that constructing the design, movement, and context surrounding future robots could lead to increased social acceptance. Similarly, choreographers leverage and create elements, e.g., a program note, to manipulate audience experience.

The background provided here, as well as the results of the work presented, in Section \ref{results}, show evidence that also contradicts this idea of universally affective motion (such as a motion profile for a particular robotic platform or artificial avatar that is ``happy'').  Moreover, in the context of dance, where expressing complex ideas in movement is the \textit{task}, the fallacy of ``expressive'' movement is also revealed.  Instead we propose an approach that recognizes the malleability of motion labeling, which is a function of culture, context, and coincidence and requires embodied expertise to navigate.  In this approach, on the side of human movers, we adopt the idea in dance where through training in many styles and techniques a particular \textit{dancer} might become, as a moving platform, more \textit{expressive}.  In robotics this is a function of hardware, software, and integrated system design.  The modifier ``expressive'' is likewise not applied to \textit{movement} but to \textit{platforms}.  This clarity in terminology is just one outcome from deep exploration in embodied methods.

\section{Methods: Embodied Practices from Choreography and Somatics}\label{methods}


%

Graduate and undergraduate students in the RAD Lab are exposed to the different aspects of choreography, improvisation, and LBMS as a tool for meaning making through recognizing patterns , reconciling paradox, contexting (meaning-making), and gaining respect for the complexity of human movement.  We approach these ideas through both functional and expressive movement experiences that can illuminate the body as basis for our ``knowing'' of the world in order to inform our engineering work.  This section highlights a small subsection of the activities in the lab with choreographic tools and often a somatic perspective, which have become important \textit{methods} for our research.  These methods are the main subject of this article, and may be foreign to many readers, and as such are presented here, rather than at the end of the article.  

\subsection{An Icebreaker That Brings Movement to the Foreground}

So-called ``name games'' are the perfect way to break the ice in any group, class, or workshop setting.  When participants know each others names, they are more at ease.  When participants are forced out of their comfort zone in a low-stakes en masse framework, they are more at ease.  Conveniently, a great way to learn people's names is through sophomoric, sing-song activities that put even the most famous roboticists' (or dancers') egos in check and embolden uncertain discourse.  Such an activity is quite important in interdisciplinary settings where participants are inherently ignorant of at least half of the expertise in the room (if not likely more), which is a fundamentally scary professional experience.  
Moreover, for a group of engineers, the topic of the workshop, movement, is invisible.  Humans move all day everyday.  So, unless one is trained to isolate and inspect it, as dancers and choreographers and somatic practitioners are, it is invisible.  Thus, an equally important goal of this activity is to make this ether visible.

The exercise is a twist on a standard, silly name game.  In the classic version of this game, participants go around the circle and say their name at the same time as making a personal movement.  The first participant (usually a facilitator) will say:  ``Hi, I'm Amy,'' and in time with their name will make a move, which can be a combination of postural and gestural change and may involve weight shift.  Typically the stakes of the room and length of the workshop can determine how involved or complex this first movement is.  It should both invite participants to move more than they are expecting, but also respect the boundaries implied by the context of the workshop.  For example, if participants are in business suits and skirts, a large jumping jack movement will only serve to alienate participants.  Typically a large gesture mixed with a more subtle postural change sets an easy tone for future participants to match -- they can exaggerate and choose a more athletic movement for their name or they can use a familiar, pedestrian gesture like a wave.

Next, the facilitator instructs the whole room to repeat their name and their movement in unison.  This is a magical moment for group building.  The facilitator sets the tone by breaking the ``movement taboo'' that can exist in engineering and now the whole group will follow suit.  It can be helpful for the facilitator to have people in the group that they know will be comfortable with the action and join in -- this will help along participants who may be resistant to such an odd request for their context.  The exercise now shifts to the next participant who will repeat the actions of the facilitator.  This time the group will repeat the facilitator and the first participants name in sequence, forcing memory and building the shared movement experience of the group.  This repeats until everyone in the room has had a turn, and a seemingly impossible feat has been accomplished: an entire room of engineers is dancing together.

If time allows, the twist on this exercise illuminates this feat further.  Now the facilitator allows each participant to go around the room in turn and correct their fellow participants in the execution of their personal movement.  To keep plausible deniability, the facilitator may not want to demonstrate, but simply choose an eager participant to go first, offering encouragement or asking questions about how the movement should be done to spur corrections.  Now the participants are engaging in iterative choreography.  Often, they'll need to work hard to articulate how a movement should be done, particularly as questions arise.  This offers the facilitator a chance to foreground taxonomies (like LBMS) which may help in this process, e.g., by using their own expertise to highlight  and articulate more nuanced differences in execution.  Eventually, this provides a point of comparison that brings the focus back to an engineering perspective: how can everyone be doing the same thing if we all have different shaped bodies?  Indeed, it is a high-level notion of movement that is needed to make such a claim, and the tools inside LBMS provide ways to improve and highlight successful replication.

\subsection{Weight, Flow, Breath, and Group Sensing}


Nearly all traditional movement classes begin with a warm-up.  For ballet, it is the barre, for jazz it is a series of isolations in the center of the room, for tap it is repetition of the core steps at decreasing intervals.  Each instructor develops variations on these warm-ups to suit the particular structure of their classes.  To break an implicit taboo around sharing a non-chaired, open space, communicating through movement, exchanging focus, and attending to the body, a three-part warm up we have utilized is described here.  This exercise has been employed in groups with no prior movement training and in sizes between about four participants to up to thirty.  Indeed in many instances only one part or a riff on one part is the right focus for a particular group, depending on the objectives of the workshop or collaboration.

Standing in a circle, the facilitator begins the movement by using an imitation and repetition strategy that invites participants to ``follow'' their movements and yet make their own unique choices in response also to the verbal descriptions of the experience that the facilitator narrates. The movement often begins by activating weight (personal) in relationship to gravity, bouncing, wiggling, jiggling, shaking and vibrating both core and limbs standing with two feet on the ground. As the movement range increases the facilitator invites the participants to activate a conscious awareness of their breath, activating flow through sounding and allowing a connection with the others in the room. Focusing on breath allows an inner/outer relationship to be established in which participants are feeling their body's inner space in relationship to the general space.  The intensity, speed, and space of the movements continue to increase with adding weight shifting from the core not only up and down, but also forward and backward and sideways. This establishes the ability to locomote and move in a shared rhythm. In doing this the group is literally ``warming up''  -- core body temperature is raised, and mobilization and oxygenation of muscles occur.  Breath support is activated and helps to establish a shared space and connection between participants.  This weight and flow sensing can be effective preparation for a shared, embodied research session.

A second exercise allows participants to go into a deeper relationship with one's own body and breath.  Beginning with a partner seated on the floor back to back, participants are instructed to close their eyes and bring their attention to their breath as they sit in relationship to another person. After several breaths and a prompt to observe their natural breathing rhythms participants are asked to bring their attention to Lengthening and Shortening in their torso, investing in the Vertical dimension with their breath. After a few breaths here, they are encouraged to place their hands on their abdomens and with each breath focus on Bulging and Hollowing in the Sagittal dimension. And finally, with hands on either side of their torso, to send breath into their side space as they invest in Widening and Narrowing in the Horizontal dimension. By attending to breath in relationship to another person and with the addition of self touch for kinesthetic and proprioceptive awareness, participants begin to sense more deeply into the shaping of their torsos and the role of breath.  

Participants are then invited to yield their weight into their partners and find their way to standing. As they begin to walk, they are asked to pay particular attention to their breath and their connection to other people in the space. As they continue to ambulate through the space, everyone is prompted to tune in and come to stillness as soon as one individual decides to stop walking. Once everyone comes to stillness, anyone may restart the group by choosing to begin walking again. The goal is to become as close to synchronization as possible. Because this is typically done with a large number of individuals in the room and they are all walking in different directions they have to rely on more than their eyes to synchronize with the group. As this exploration progresses, attention is brought to the length of the pauses and whether or not predictable rhythms are developing. Eventually participants are prompted to set the pace of the locomotion between stops so that whoever starts after a pause determines whether or not it is a fast paced run, a slow, cautious walk or somewhere in between. 

Another short exercise can further develop the sense of the group as a whole, bringing a sense of play as well.  One participant, a volunteer, exchanges eye contact with the participant to their left.  Using this eye contact as a nonverbal line of communication, the pair attempts to clap at the same time.  This is trickier than it sounds because it is a tendency to lose eye contact, which can feel culturally awkward, initially, and undervaluing the complexity of simultaneous action, participants tend not to give the task their full attention.  However, after passing this around the circle a few times, most participants receive the satisfaction of simultaneous action, emphasized with the sound of clapping, and the energy in the room will invariably change.  The participants are ready for more.  Next, to increase the complexity of the task, participants are instructed to add the possibility of locking focus with others across the circle and increasing speed of the exercise.  
This warmup builds shared focus, group sensing, prolonged awareness into the exercise (you never know when the focus will be directed to you).
It also demonstrates how synchrony in dance is temporal, spatial, and qualitative. Dancers who synchronize well attend to many factors, can observe them, and have a range of choices in their body movement repertoire to adapt, adjust, and, thus, synchronize. This training is additive and not reductive.  The abstraction of synchrony (it's an abstraction, not an absolute truth as in same physical behavior) needs additive choice in order to exist on the multiplicity of morphologies of human bodies.


\subsection{Kinesphere Exploration Through Spatial Pulls}\label{spatialpulls} 

Laban defined the Kinesphere as ``…the sphere around the body whose periphery can be reached by easily extended limbs without stepping away form that place which is the point of support when standing on one foot, which we shall call the stance/place. We are able to outline the boundary of the imaginary sphere with our feet as well as with our hands…when we move out of the limits of our original kinesphere we create a new stance…We never, of course, leave our movement sphere but carry it with us always, like an aura'' \cite{von1966language}.
This idea includes more than a roboticist's notion of workspace, but it is similar.  There are many ways to explore the dynamic space around which humans move. 
Primarily, in the study of the Kinesphere, LBMS maps an approach to Kinesphere that includes Zones, Levels, Reach Space, Pathways, Forms, and Directions.  Patterns of Body Organization which are underlying movement pattern identification based on human motor development that support relationship of part to whole provide another way to internally organize movement within the Kinesphere. ``Our form, with its upright vertical stance, and bilateral symmetry, is organized through relationships among upper and lower body, right and left sides of the body, front and back, core and periphery'' \cite{studd2013}.

One exercise in particular highlights spatial organization keyed around a room versus a body reference frame -- exploration through Spatial Pulls.
To explore the Kinesphere through Spatial Pulls, participants, moving around in a dance studio, are instructed to move while attending to specific directions within the room, e.g., ``Place High'', ``Place Middle'', and ``Place Low''.  Initially, these directions are the same for every participant, keyed off the room, and using the hand, or another body part available to all participants, comfortably extended, as the body part drawn to each direction.  Gradually, the notion of a space centered around each mover individually, allows for instructions keyed from each participants body, e.g., ``Forward Right'', which will result in more varied movement for participants with different facings.  Another changing condition can be the extent to which the hand moves into the space: a shorter extension is in Near-Reach Space, while, conversely, full extension of the hand is in Far-Reach Space, which can also invite locomotion.  Finally, the body part which initiates the movement can change:  elbow, hip, top of the head, nose, etc.  Inviting participants to debrief the exercise, e.g., answering ``When do you do these actions in your daily life? or ``How did different spatial prompts make you feel?'', reveals commonalities and differences in experience that highlight the complexity and malleability of meaning-making through movement.

These changing conditions highlight several lessons:  1) the body is an adaptable and accommodating phenomenon to changing requests, 2) what delineates a particular body part is not always clear, and 3) space has meaning.   This framework abstracts away the physical properties of the moving parts, focusing on high-level movement ideas common to all movers. ``Place High'' is a different place for every mover based on different heights, shoulder flexibility, and ability to rise onto the ball of the foot.  Moreover, each mover has different prior experiences -- maybe positive or negative or something complexly in between -- with this kind of movement, which may create strong or subtle emotional responses to the components of the activity.  That is why, for understanding this framework, one cannot just rely on going through the related literature that covers these details. Instead, one has to get inside a movement studio, experience the different spatial pulls, move through different spatial levels, and imagine the different spatial instructions while moving around. 


\subsection{Written and Embodied Movement Observation}
Writing about movement can be an easy inroad to exposing engineering students to more complex movement phrases.  Moreover, this kind of descriptive writing is very good experience for engineers who need to learn how to describe complex systems as part of their work.  A short exercise in such an observation might simply involve bringing up a YouTube video of some movement and having participants write about the video, describing it to someone who has never seen it.  A sentence might read as follows (this one was written about Mikhail Baryshnikov in Twyla Tharp's ``Push Comes to Shove''):  \textit{The dancer enters a foggy spotlight emerging from shadows upstage.  He deftly flips a tan hat into the air and pops it on his head with a nonchalant air. Holding his hands limp, softly locked in a spatial location while creating a myriad of choices with his core underneath.}  This description requires participants to watch movement in great detail and make sense of the pattern.  To take the exercise a level deeper, participants can write about distinct movement examples and switch samples to see if a naive reader can determine what the original movement sequence entailed.

To continue to work on participants listening and movement observation skills -- from an embodied perspective -- participants are placed in pairs and prompted to engage in what is often referred to as a mirroring exercise. With partners facing each other, one is directed to improvise various movements which can include pedestrian movements, weight shifts, and arm and leg movements, while their partner is working real time to mirror/imitate the exact movements that they are observing. This tool is used to hone movement observation skills and the ability to make movement choices in the moment.  Replicating exactly is not possible -- because both participants have different bodies (different ranges of motion different lengths of bones), but we find that experience in this kind of exercise can lead to richer interactions and imitations.

\subsection{A Notation System For Movement Analysis}
\begin{figure}[h!]
\centering
\includegraphics[width=.95\columnwidth]{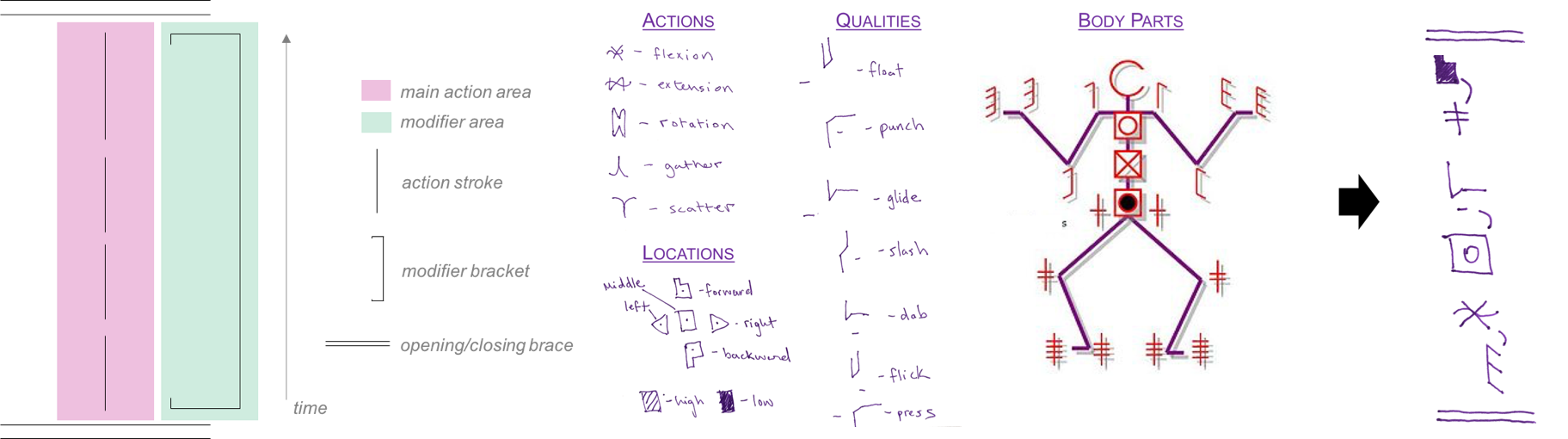}
\caption{The diagram at left shows the structure of a vertical Motif (shown at right).  The center column of a Motif denotes the main action. The length of a stroke indicates relative duration. These generic strokes may be replaced with more complex symbols (shown in center). The green area to the right, which may also be called a theme bow, allows for the main action to be modified, possibly highlighting overarching themes in the phrase. The sequence of action moves from bottom to top. }
\label{motif}
\end{figure}

Labanotation is a well-known toolset that utilizes Laban's work to record movement sequences -- mostly for archival purposes.  A more common shorthand notation system is called Motif, which is useful in practice for applying (and learning) the tools in LBMS.
While Labanotation is a precise record of a specific movement phrase, recorded for posterity, Motif is a description of a gross movement idea or pattern. Each form utilizes its symbols in slightly different ways. Utilizing these notation schemes can clarify observations and record executions of movement. Something  evident from study of LBMS is that the phenomenon of human movement is rich, its perception is rich, and singular records or interpretations are unlikely to universally categorize movements (e.g., in the gesture-based Nest system where a waving arm is both a sign of no danger and danger \cite{nest2}). Thus, our group favors use of Motif over Labanoation.  The tool of Motif helps designers consider which features or broad patterns of a movement are most important for a given context; the mechanics of the tool are described through the diagrams in Figure \ref{motif}.

\subsection{External Quality and Internal Motivation in Movement}
Another useful activity revolves around the LBMS component of Effort and reading/interpreting Effort Motifs. Effort can be simply described as the ``quality'' of a movement --  the mover’s inner attitude, motivation, and intention as revealed in observable movement.   Activities designed to allow participants to explore the choices in this space can evolve as follows:
\begin{enumerate}
\item Define and discuss the idea of Effort, an important element of the theory explicated in the Laban/Bartenieff Movement System \cite{laban1947effort,studd2013}
\item Engage in free movement improvisation in response to the stimuli of varying musical selections
\item Engage in free movement improvisation in response to words (adjectives) that evoke qualitative movement responses
\item Playing children’s games, like Red Rover, Ring-Around-The-Rosey, and Duck-Duck-Goose, which naturally elicit playful, dynamic motion profiles (which easily produce short moments of complex Effort constellations, such as States and Drives, providing fodder for discussion)
\item Break into groups that are each given an Effort Motif and have the group ``decode'' it into movement that falls on the Functional end of the F/E Theme
\item A second group has same Motif but goal is to generate a movement that falls on the Expressive end of the F/E Theme
\item Show movement sequences that are generated from the same Motif side-by-side. What is essential about the movements and their commonalities? What makes the sequences different and ``mean'' different things? What are the contextual cues being employed by each group?
\end{enumerate}
An exercise like this can be led by an CMA or CLMA who can use their expertise to highlight and describe moments of interesting Effort patterns that emerge from these exercises.

\subsection{Establishing the Malleability of Meaning in Movement}

 
 Movement as narrative or storytelling is well evidenced in classical ballets like Swan Lake or Don Quixote.  In those instances, choreographers are committed to revealing specific narrative elements.  This scale of ``meaning'' -- whether narrative (``Romeo loves Juliet''), emotional (``that part seemed sad''), or aesthetic (``that reminds me of a flower blooming'') -- is utilized to the choreographer’s discretion and this utility impacts the audience’s subjective experience and interpretation.  To illustrate this, an exercise described here allows students to create mini-narratives and manipulate contexts from the pedestrian (``a shopping trip to the mall'') to the surreal (``swimming through Saturn's rings'') with a single gesture.  
 
 All participants sit against one wall of a large, open studio.  A facilitator selects a random\footnote{A good facilitator of group dynamics rarely makes a random choice; selecting an outgoing yet serious participant can be an effective choice here.} person to begin the exercise.  The person is then directed to ``pick a scene, place, idea, event, dream, etc.  It can
 be an atom floating across Saturn’s moons, it can be wind, it can be a dragon lying in an abandoned mythical forest, it can be a monk sitting atop a hill.  There are no limitations to the who, what, where, when, why.  Decide what or who you will be within this scene.  Are you the heat coming off a loaf of fresh bread?  Are you Cleopatra on her throne?  Now when you stand up and enter the large open area of the space, you will be this.  No speaking or verbal clues.  You will move like it and commit to it.  Go.''
Now this first participant, the ``scene leader'' goes into the space without revealing their internal image or desired role; they simply begin moving the space.  After the scene leader begins, the facilitator selects students at random and instructs them to, ``Decide what you think this is.  What scene/what action/what is happening?  What space is being created?  And then select
 how you will insert yourself into this.''  This participant now enters the space, similarly without revealing what they think is going on or talking in anyway; they simply begin moving, in earnest, their role.  This continues for all of the participants until a dynamic, moving scene has been created.
 
In one example, the scene leader laid on their stomach with arms outstretched, holding an imaginary item, and keeping one eye closed.  They rolled back and forth on their stomach with constrained breathing and bound overall flow.  It was revealed later that they were a World War I soldier fighting in the trenches.  Without this knowledge, other members of the class cohort decided for themselves what was happening and joined the scene.  The first ``joiner'' began moving in a curved path, on their hands and feet with hips in the air, similar to a children’s camp ``bear walk''.  They would occasionally pause and switch their focus
 up from side to side.  This was a labored movement pattern, with dramatic weight shifting and a strong energy.  It was revealed later that the first joiner believed the scene to be a safari and inserted themselves as a large hunter-target.  
 
In each case, the participants are told, ``once you join, continue what you are doing until the exercise ends''.   At the end of the exercise, each participant had joined the scene and committed to their interpretation.  After a few minutes, the facilitator ends the exercise and participants stop moving, and then each individual revealed what they believed the scene to be.  The interpretations of one single scene can be varied: from ``gas along Saturn’s rings'' to ``children on Easter'' or rather unified: ``journalist reporting on war'', ``soldier patrolling an area'', and ``medic caring for the wounded''.  Participants altered between joining early in the scene and later in the scene.  Additionally, as a further variation, a subset of participants can exit a scene once it had been ``finalized'', and then asked to reinterpret the scene from their original image. This creates a cycle of constantly deciding and participating based on the new visual representation.  

The lessons learned can be numerous.  For example, one person is a ``hunter'' when another is the ``buffalo''.  How does this
 change when the hunter disappears?  Relationships created through movement and space are altered through perspectives, additions, and subtractions.  Confirmation bias exists in performance as audience members process what may be purposefully ambiguous or unstructured. 
 One scene leader might choose a very specific theatrical action, like the scene leader on their stomach, while the joiner may use an entirely improvisational, constant, and freeform movement pattern.  This autonomy for the joiner can warp the scene leader’s
 vision and give an entirely new scene idea to the audience, or they may choose to best interpret the scene leader’s decision and support it.  This distinction was critical and constantly re-emphasized by the facilitator: ``Do not simply decide what the scene
 leader wants and add yourself to their vision.  Decide what you think it is
and how you will insert yourself into that scene.''  This exercise illustrated the malleability of meaning in movement.  In real time, participants can observe
 how movement patterns reorganize into new narratives, for individual performers and for the performance area as a whole.

\subsection{Building Shared Values} \label{values} 


\begin{quote}
\textit{This class will be unlike any class you have taken before. You do not necessarily need to be a dancer or an engineer to successfully participate in this course. However, as we forge our way into unfamiliar territory, investigating the commonalities and differences between two disciplines, we ask that you bring your adventurous, inquisitive selves to class. By the end of our time together you will be able to illustrate concepts from dance and engineering, and demonstrate the interplay between these disciplines. We will move, physically embodying basic concepts from somatic theory; program, deciding discrete allowable orderings of actions for automated systems; and work together as a group to create an environment where these ways of thinking about movement (both quantitative and qualitative) can coexist. We will be doing readings designed to facilitate broader thinking, group assignments involving movement generation and observation, and exploring the inter-workings of automated systems.}
\end{quote}

A dance class is a different environment than an engineering class, and a major challenge in such a course or workshop is ensuring students and teachers or participants and facilitators feel safe and respected within the time and space.  For example, consider the infrastructure alone: the former is a wide open room, often with large mirrors, where students wear leotards or loose-fitting clothing that moves with their skin; the latter is a chaired environment, often in a stadium-style arrangement, where students take notes in street-style clothes.  Moreover, conventions around student-teacher contact, lateness, and expected pedagogy can differ vastly.  The above description is an excerpt from  DAN 3559 / ENG 3501 Electronic Identity and Embodied Technology Atelier, cross-listed between the dance and engineering programs at the University of Virginia in Fall 2014.  It's the kind of text that helps to bring students from these two communities together in a shared space.

Physical movement can often be intimidating, memories of evaluative movement (running a P.E. class sprint) or socially acceptable movement (appearing ``cool'' at a school dance) might render a new participant wary of embodied exercises.  Teachers might be accustomed to working with students who have had years of exposure to various dance techniques -- thus posing challenges of where to begin and how to earn committed investment from the students while mitigating the risk of student rejection.  A set of governing principles around these exercises can counteract many of these presumptions and establish a safe space from the beginning.  Ground rules need phrases like, ``we show support by being open minded'' and ``this is a safe space for your body, free of judgment from yourself and your peers''.  Following from this practice, in our experience most all participants present for each exercise participated fully in each exercise.  When common ground rules are enforced, no one has ever excused themselves from an exercise or rejected the exercise itself.  Allowing each participant to engage in the way that fits their personal comfort level is an important way in which we celebrate and facilitate the vastness of human motion in these exercises.

Another challenge is demonstrating mutual respect for values, as engineering is often labeled as a ``quantitative'' field, while dance is labeled ``qualitative'', with a resulting immediate division inside such an interdisciplinary collaboration.  Engineers might be tempted to look down on those unable to code or lacking depth of mathematical training; dancers might be tempted to call engineers ``linear thinkers'' who oversimplify important phenomena.  Facilitators can emphasize the places where these alternative methods are most powerful, giving respect to both groups in the room.  Participants often found articulating their response to movement or creating movement to be a detailed, semi-scientific process with need for hypothesis, evidence, and synthesis.  Additionally, quantitative thinking requires creativity and interpretation.  Crystalline, unequivocal answers are rare despite troves of quantitative evidence.  A more nuanced quantitative/qualitative contrast is evidenced in embodied experience.  For example, doing a movement from a ``quantitative''
 impetus, such as five repetitions or 60 seconds, versus doing a movement from a ``qualitative'' impetus, such as a ``surprised expression''.  Both such prompts produce objective and subjective differences in movement phrases created in response.

\section{Results: Progress in the Components of Expressive Robotics}\label{results}

The methods described above have been utilized in a myriad of robotic projects, in the RAD Lab and outside it.   Moreover, in successful funding proposals, these methods have been explicitly touted for their potential for dramatic leaps forward inside robotic control and human-robot interaction.  Here, we highlight specific projects and how these approaches have informed the work within. 

\subsection{The Role of Context in Resolving Expressive Movement}\label{madi_sec}

The affect experienced by a human viewing a stylized movement, either human and robotic, is heavily dependent on the situation where the movement is being experienced or imagined. If no situational or environmental context is provided, there is a significant chance that the observer will misinterpret the affective intent of the movement \cite{zeng2009survey}. For example, imagine a human walking, stomping forcefully, and thrashing their arms side to side. If this human were in a classroom, this movement demonstration may be viewed as anger or frustration brought on by a bad score on some sort of assignment. However, if this human were in the thick underbrush of a jungle, the same movement profile may be necessary for the human to navigate through the terrain and may not be indicative of any particular emotion or feeling. 
The previous example, and examples like this that were discussed among members of the RAD Lab, were inspiration for a series of user studies that were constructed to investigate the role of environmental context in affect recognition. The user studies used stimuli included images from the OASIS database \cite{kurdi2017introducing}, stylized walking sequences produced by animators in \cite{etemad2016expert}, and the stylized walking sequences from \cite{etemad2016expert} superimposed onto videos from YouTube that were similar to selected images from OASIS. 

The primary goal of the first study was to investigate how priming a human with an affect label influenced the environmental context where the walking sequence was envisioned. The study presented 3 sets of 7 OASIS images to participants accompanied with a labeled stylized walking sequence (feminine, masculine, happy, sad, energetic, tired, or neutral as created by expert animators in  \cite{etemad2016expert} ) on a white background. The participants were asked to select one image from each image set where they envisioned the pre-labeled walking sequence occurring. The results from the first study were expected in the sense that environments such as the beach or a scenic nature view were often associated with the feminine and happy walking sequences while scenes of war or destruction were associated with the masculine, sad, and tired walking sequences.

The primary goal of the second study was to investigate the effects of environmental context on human perception of the affect displayed through stylized movement when no affect label is provided to the human. The participants were presented with 56 walking animations (combinations of 7 walking sequences and 7 environmental contexts as well as white background walking sequences) and asked them to label each animation as either feminine, masculine, happy, sad, energetic, tired, and neutral.  At least two observations are immediately evident when interpreting the results presented in \cite{madi2017ICSR}: (1) people seldom interpreted the affect displayed through the animations as the affect originally intended by the animators and (2) the affect label selection rate was influenced by the environmental context. 

For example, the selection rates for the labels that match the animators’ intent are relatively constant for all of the stylized walking sequences displayed in front of a garbage-filled scene, but the majority of the selections that contradict the animators’ intent were sad, neutral, and masculine, indicating a bias in viewer choice. As in Study 1, the Beach scene was often selected as happy while the War and Garbage scenes were interpreted as masculine and sad. However, the results from the Aspen, City, and Lightning scenes in this study were not at all comparable to the results from Study 1, indicating that the walking sequences may have had a significant impact the affect selection. 
The first and second study confirmed, at a high-level, that environmental context is a critical component in affect recognition of stylized walking animations, but provided little insight into the individual contributions associated with the affective ratings of the environmental context and the stylized walking sequences. Future studies will dig into this phenomena further using the semantic differential rating scales presented in \cite{bradley1994measuring}.

\begin{figure}[h!]
\centering
\vspace{-.1in}
\includegraphics[width=.95\columnwidth]{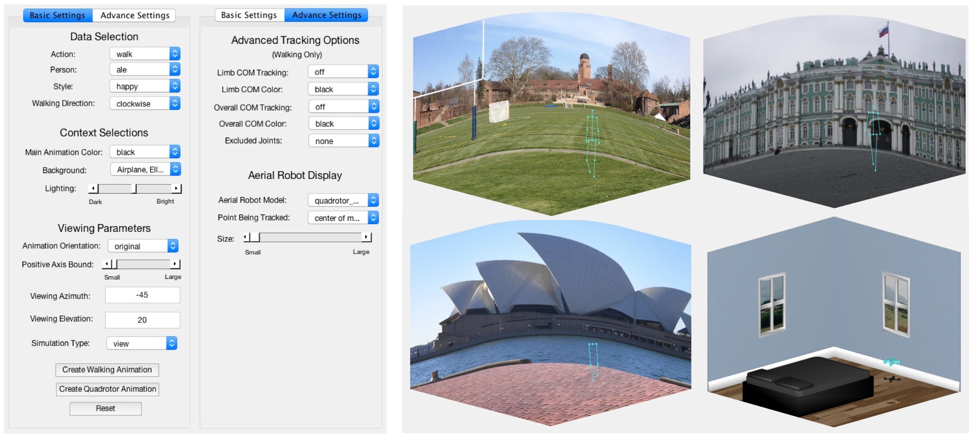}
\caption{This figure contains a screen capture of one of the MATLAB Tools developed in the RAD Lab (left) to test the effect of context -- as approximated by background image and shown in created stimuli (right) -- on motion meaning.  The animation screen captures at right include a human walking at Cranbrook’s Upper Fields in Bloomfield Hills, MI (top left), a human without knees and elbows walking at the Winter Palace in St. Petersburg, Russia (top right), the lower body of a human walking at the Sydney Opera House in Australia (bottom left), and an aerial robot traversing across a virtual bedroom (bottom right).}
\label{madi2}
\end{figure}

The user studies documented in this section are a preliminary step towards understanding the relationship between movement, context, and affect recognition. We are also developing MATLAB tools that can be used to quickly and efficiently create and test stylized trajectories in various environmental contexts using data-driven approaches. Currently, the data that is being used to create the trajectories is stylized walking data obtained from an online database at the University of Glasgow \cite{ma2006motion}. However, in the future, the tools could be used as platforms for simulating choreographed movements in applicable HRI settings to quickly gauge if the movement profiles are received well and are adequate for a desired application. A screen capture of one of the current simulations is shown in Figure \ref{madi2}.

\subsection{Improving Creative Workflow for Designers of Robotic Motion}\label{alli_sec}

\begin{figure}[h!]
\centering
\vspace{-.1in}
\includegraphics[width=.85\columnwidth]{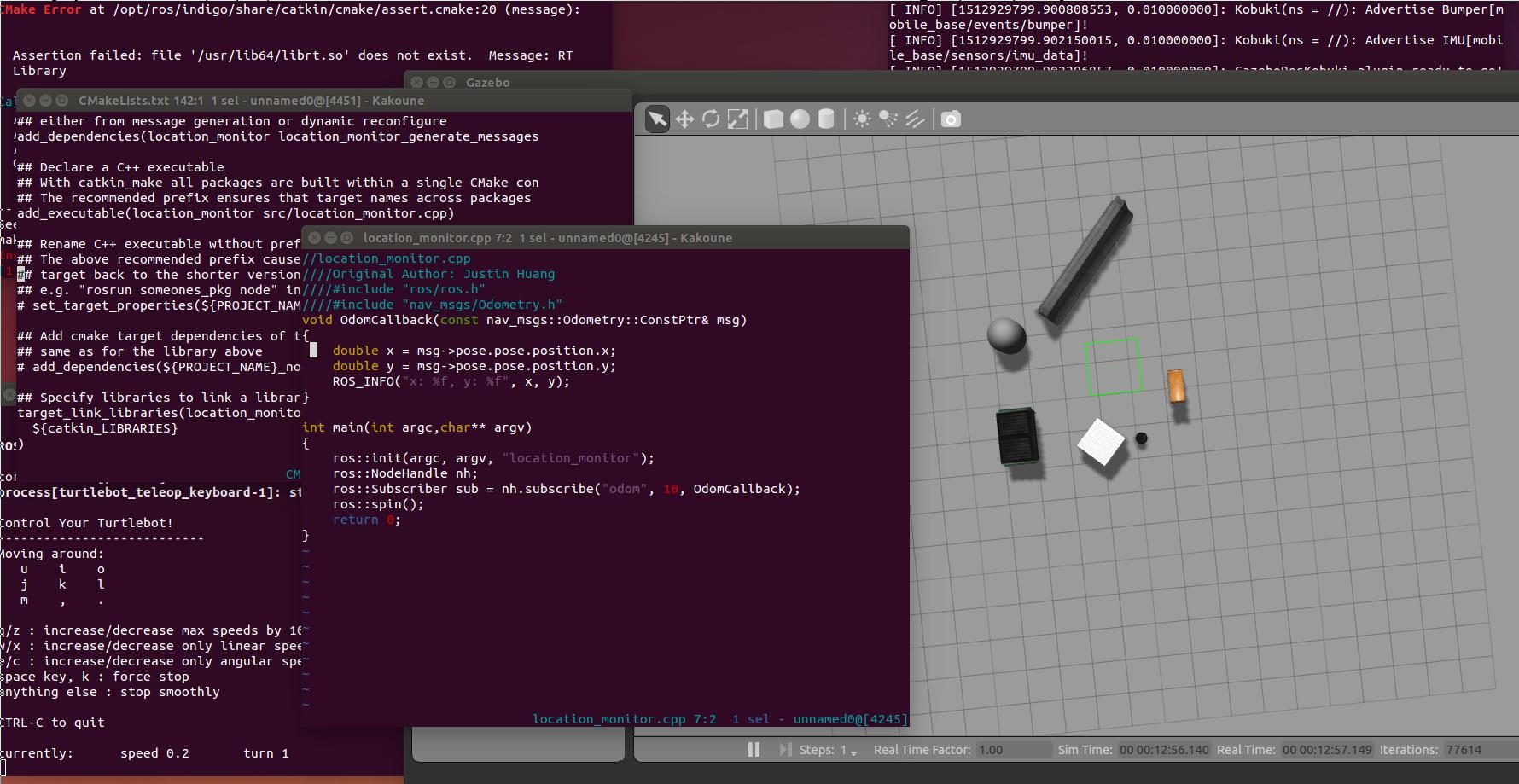}
\caption{A typical ROS interface, featuring multiple command-line interfaces
and a physics-based simulator. Figure from ``An Introduction to Robot Operating
System," an online tutorial aimed at absolute beginners.}
\label{alli1}
\end{figure}

An increasingly ubiquitous tool often used for controlling robotic systems is
the Robot Operating System (ROS). ROS is an open-source, Linux-based software framework which
standardizes common tasks in robotics, such as collecting sensor data and controlling
robotic hardware. Figure \ref{alli1} shows a typical example of the interface:
programs are developed with heavy use of the command line and simulations before
code is deployed to actual robots. As a student (jokingly) put it: ``ROS is software with the sole purpose of
forcing the user to open up many windows. It also has robot applications.''
Thus, even for technical participants, the user experience with ROS is
notoriously bad. People trying to learn the software for the first time often
quit because of how frustrating and tiresome the process is.

When using ROS to test robot motion strategies and controllers, there are many
points in the process where the user has to stop and switch the mode of
interaction with the system. First, a text editor may be used to edit code.
Then, another window is needed to stop the currently running program, reload the
code, wait for initialization, and click to another window to view the results
on the simulator. Moreover, the user may have several different
``nodes'' running, for example, one for collecting camera data and one for controlling 
a robot arm. Each of these requires its own window for editing and monitoring.
Several factors here are intimidating: the requirement to use the command line,
the difficulty of keeping track of all the windows, and the constant mode
switching between writing code and manually managing the programs.

Working with choreographers and dance artists drives home the need to streamline
this software. Artists are used to creative processes where the ``time to compile'' can be nearly
instant. Computer-dependent workflows, on the other hand, are often hampered by
unfriendly interfaces and long compile times.
Software limitations influence how people design projects, and
thus limit scientific exploration. One reason for this is that developers of software for robotics are often
people who do not have a clear picture of the skills and needs of
non-super-users. 

Interdisciplinary collaboration helps solve this problem by pulling us out of our bubble of
experience. After technical education, skills such as writing wrapper
scripts, parsing and generating low-level code, and reading through pages of
program output to debug errors, become second nature. Collaborating with people
from different backgrounds challenges assumptions about what is normal, and
highlights areas where creative workflow can be improved.

To reduce these barriers, for technical and nontechnical alike, a project that
aims to create a live-coding interface for ROS was initiated. Live-coding is a
performing art especially prevalent in computer music, where many programming
languages and interfaces exist for creating improvised music using sound samples
and a computer. Similarly, this project is creating a domain-specific language
(DSL) which is immediately compiled to a process which publishes ROS messages to
a robot or simulator. Thus, as the user edits code, they immediately see changes
on the robot with one press of a key on their keyboard. Apart from minimizing human suffering, why
decrease barriers to learning, and decrease compile time while working? The goal
is flow: the mental state of complete absorption in an activity. When engaged in
some creative process -- coding, dancing, writing, anything -- the best work is
done when the creator is focused and feels agency over the tools they are using. If we draw an (imperfect!)
analogy: the human as choreographer, robot as dancer, imagine how difficult it
would be to work with a dancer who randomly freezes, refuses to listen to
commands, who only speaks one specialized language that takes weeks of dedicated
practice to learn. Obviously, any human endeavor involving robots would be much
more productive if the creative process was able to flow more smoothly. The tool 
we are building to this end, called \textit{Improv}, is currently usable for simple 
mobile robots and the code is available online \footnote{https://github.com/alexandroid000/improv}.
User studies are being planned to measure the impact of fewer mode-switches and more 
instant feedback on the user's creative process.

This project can be a jumping off point for formalizing other identified
choreographic ``technologies.'' These technologies are tools that
choreographers use to analyze and create movement sequences. These include
notation, motif, and video recording, as well as creative tools such as
arrangement of movements in space and time. Movements can be reversed,
retrograded, or reflected across different axes. Similar movements can be
performed at different extents or levels, or at different tempos, or with
sharper or more gradual speed changes. Moreover, dancers and choreographers have
identified heuristics that explain how these different technologies tend to
affect how audiences interpret dances: why we tend to see some dances as
frantic, others as peaceful, and what techniques create tension or capture
attention. By automating these heuristics, it may someday be possible to program
a robot at a much higher level than is currently done: the instructions from the
human to the robot will be more similar to those of a choreographer. We are far
from this point today, but by collaborating with artists and movement
observation experts, we can accelerate toward this goal.

\subsection{Motif as a Source for High-Level Movement Abstractions}
In contexts such as teleoperation, robot reprogramming, human-robot-interaction, and neural prosthetics, conveying movement commands to a robotic platform is often a limiting factor. Currently, many applications rely on joint-angle-by-joint-angle prescriptions.  This inherently requires a large number of parameters to be specified by the user that scales with the  number of degrees of freedom on a platform, creating high bandwidth requirements for interfaces.  We have developed an efficient representation of high-level, spatial commands that specifies many joint angles with relatively few parameters based on a spatial architecture, appropriate for quick coarse positioning of teleoperated platforms.
Key inspiration for platform-invariance in spatial commands for robots comes from LBMS. The relationship between Body and Space in LBMS provides terminologies for movements that can be applied to different human morphologies irrespective of the differences in physical features and forms (indeed, all humans have at least slightly different morphologies).   Starting explicitly in an embodied place helps to bring these approximations to the forefront.

\begin{figure}[h!]
\centering
\begin{subfigure}[b]{.25\textwidth}
\includegraphics[width=\textwidth]{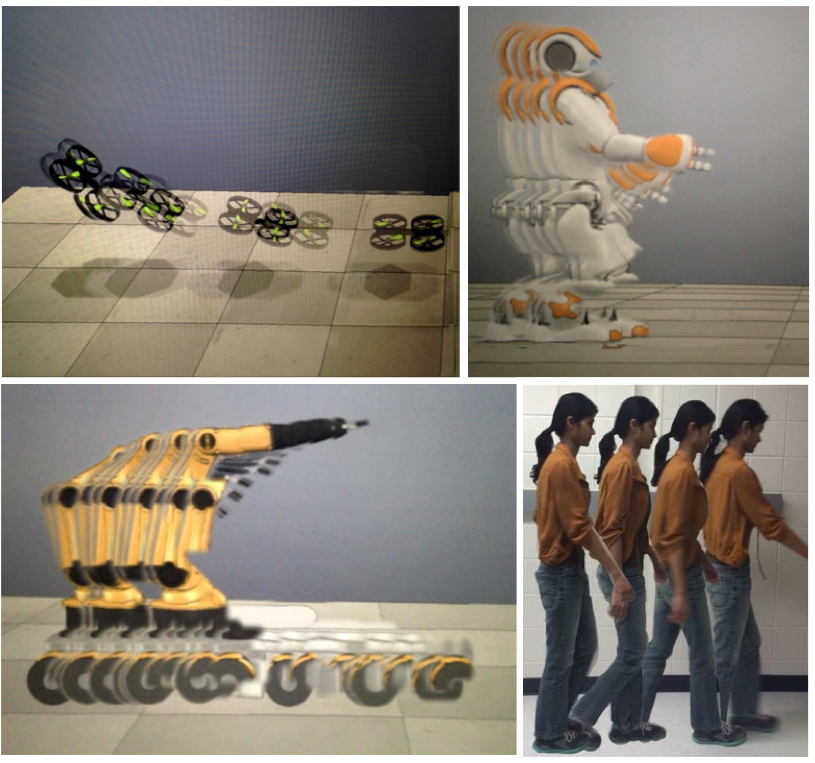}
\label{forward}
\end{subfigure}
~
\begin{subfigure}[b]{.3\textwidth}
\includegraphics[width=\textwidth]{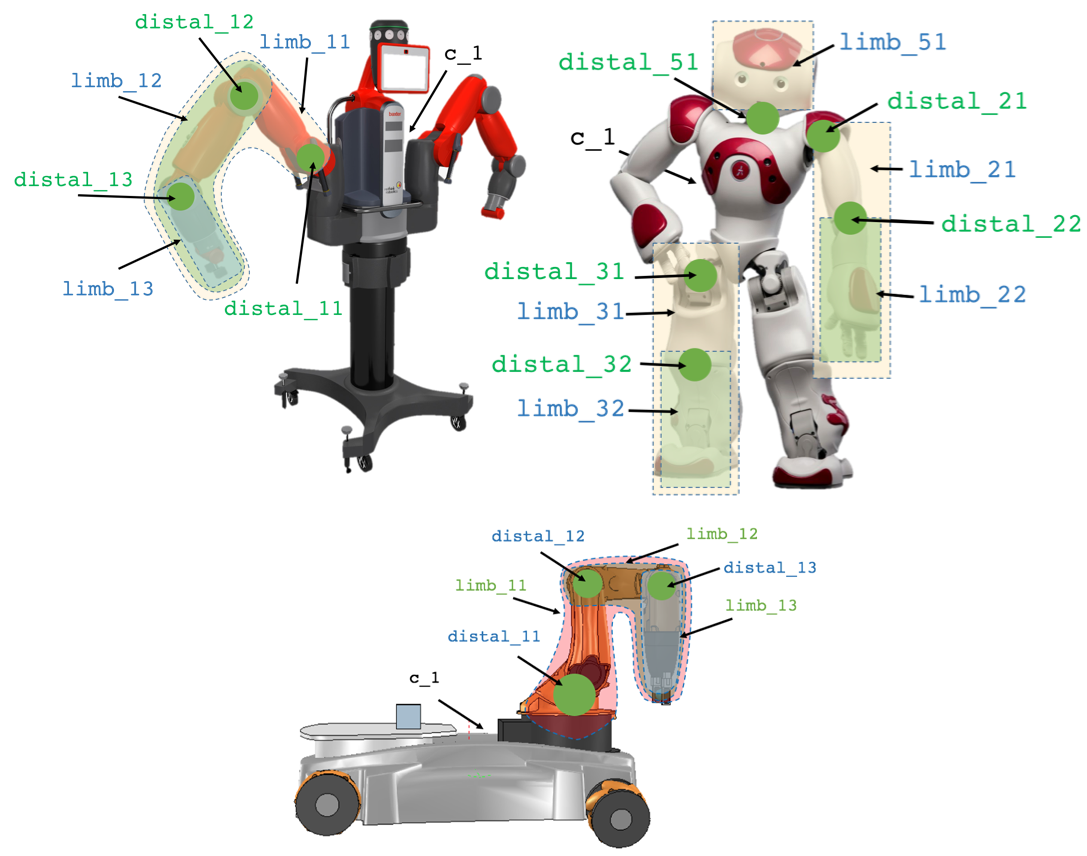}
\label{labels}
\end{subfigure}
~
\begin{subfigure}[b]{.37\textwidth}
\includegraphics[width=\textwidth]{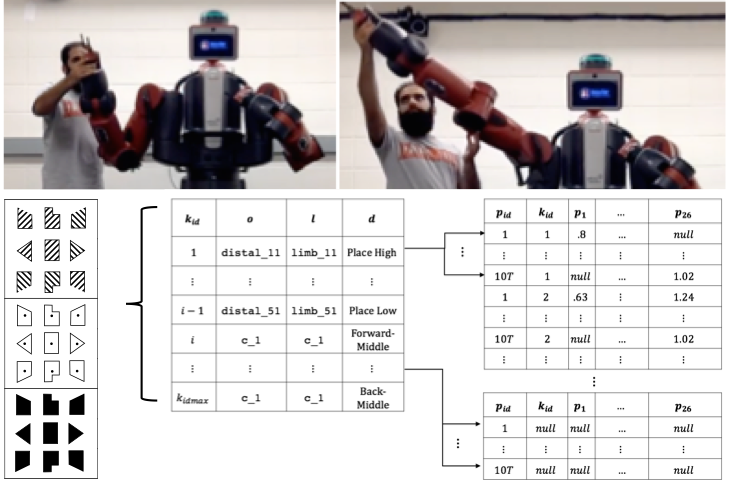}
\label{training}
\end{subfigure}
\vspace{-.1in}
\caption{Left: The notion of ``forward'' on distinct platforms.  Center: Overlapping label sets leveraged in generating ``the same'' behavior on distinct platforms.  Right: Manipulating a platform through a ``scale'' that instantiates a database of poses, indexed by high-level symbols.}
\label{ecl}
\vspace{-.1in}
\end{figure} 

An overview of the general method for labeling connected platform linkages, generating a databank of user-specified poses, and mapping between high-level spatial commands and specific platform static configurations is given in Figure \ref{ecl}.  This architecture is ``platform-invariant'' where the same high-level, spatial command can have meaning on any platform.  This has the advantage that our commands have meaning for human movers as well.  The overlapping, redundant labeling scheme used, termed Expressive URDF (eURDF), allows for the operator to zoom in on broad areas of importance on a platform (for example `arm' versus `forearm').  A database of poses, termed an Embodied Configuration Library (ECL), is generated by a user to translate spatial commands to articulated positions of the robotic platform.  Symbols from the LBMS Motif system, from the categories of Space and Body, index this database and form the basis of the command structure.

The final architecture was implemented for 26 spatial directions on a Rethink Baxter, an Aldebaran NAO, and a KUKA youBot. Users were trained in the Space component of LBMS, for only ~30 minutes, learning the related Motif symbols.  They then created simple movement phrases motivated by a context like ``a restaurant hostess'' and used the Motif symbols to describe their sequences.  The Motifs were corrected by an expert user (with ~10 hours of training) and then used to automatically generate robotic movement sequences on the three platforms.  The recorded human sequences were compared to the robotic platform performances by another group of users.  For Baxter and NAO, in 8 of 9 of user sequences, a majority responded that the robotic recreation was `the same' as the human movement; for the YouBot, which is not anthropormorphic, 2 out of 9 were classified this way \cite{darpaJournal}.  

The embodied movement activities used in this research provided an easy, almost rapid-prototyping, ground for envisioning the Kinespheres for different robotic platforms without having to go through extensive testing or simulation. Developing a plan for labeling different poses without having to do a trial-and-error method in code saves a significant amount of time.  When working with a two-armed platform, exercises like those describe in Section \ref{spatialpulls} help researchers mirror the platform with their own body and form a mental model for how robots are ``installed'' into this system: by moving through a movement ``scale'' \cite{laban1966choreutics}. Even when the challenge arose to label a one-armed platform, the extensive study of how our own bodies might adapt to a similar challenge saved time in formulating motion control algorithms for such a platform.

\subsection{Embodied Movement Strategies in Locomotion}\label{umer_sec}

For a typical robot design problem, the normal route is to investigate the low-level details about the platform's movement, such as the behavior of individual actuators.  In this process a complex motion profile for the whole system is defined through the behavior of small parts first. For example, in making a legged robot walk some distance, a mathematical model for the individual joints and links is needed, and then a motion profile has to be derived for these individual elements to give a whole body movement that satisfies walking.  Inspired from investigations like those described in Section \ref{methods}, we explored different locomotion themes through an embodied starting place in order to design a weight shift/locomotion system that would allow a robot to move from the ``core'' vs. through actuators on the ankles. 
This desire to more accurately replicate human locomotor patterns is not only about expressivity, but it is in fact also about efficiency. 

As described in \cite{bartenieff1980body}, human walking is composed of three high-level movements: Thigh Lift, Forward Pelvic Shift and Lateral Pelvic Shift (word ‘core’ is also used interchangeably with pelvis). The engineering team was interested in seeing if they could design a weight shift/locomotion system that would allow a robot to move from the ``core'' vs. through actuators on the ankles. It was desired to explore the many mappings of the Body and Space to understand the movement patterns and ``translate'' or ``create'' a similar pattern in a robot. It was expected that this approach could lead to more efficient gait as well as a more expressive robot overall.

\begin{figure}[h!]
\centering
\vspace{-.1in}
\includegraphics[width=.98\columnwidth]{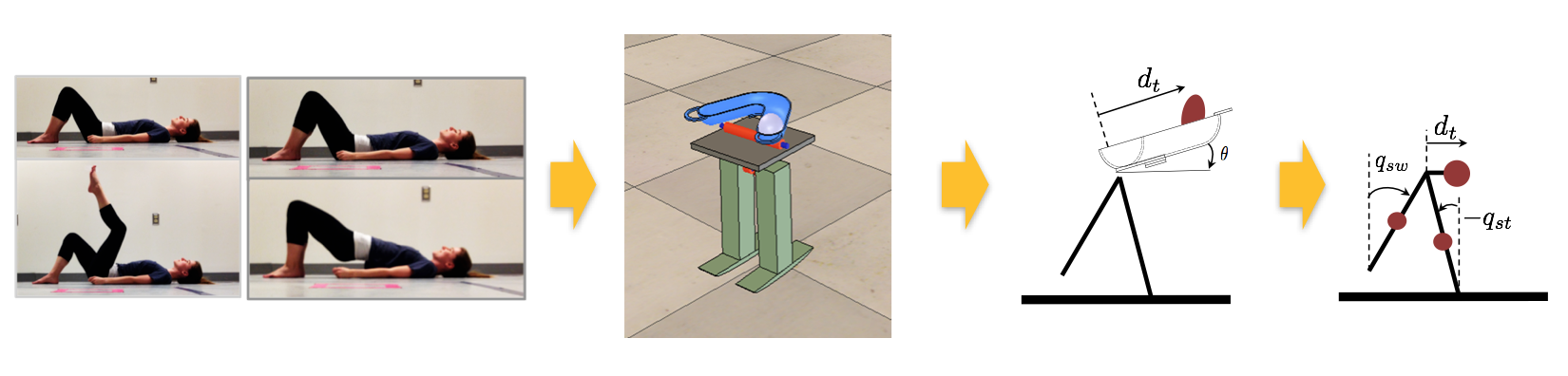}
\caption{Top-down design process: from embodied exercises practice by researchers (left) to a proposed bipedal robotic platform design (middle) to a simplified model, with only three degrees of freedom, for mathematical modeling (right). \cite{umer2016lbms,umer2016control}}
\label{walker}
\end{figure}

To help with a thorough understanding of these embodied ideas, a collaborator with certification in movement analysis brought in her experience from the Body Component of LBMS and specifically through the practice of Bartenieff Fundamentals, a set of principles and movement sequences that seeks to optimize the supportive relationship between body organization and movement intention. 
The design team was taken through some activities that involved changing relationship to gravity by working on the floor, locomoting through the space, studying skeletal anatomy and working through an experiential anatomy approach. Gaining an insight into the human movement capacity was helpful in engineering of a biped robot with core-located actuation. This collaboration has continued in the subsequent stages of the project as well.

We came up with an initial design that consisted of a tray structure with a channel for a heavy ball of mass to roll over it. The idea of using this tray was to allow the ball to move both in the sagittal and the lateral direction just as humans shift their own centers of mass in the sagittal and lateral directions. This tray design has a U-shape with length and width of this ‘U’ being the design variables. This tray structure sits on top of the two legs in the robot. Using the idea of a rolling ball to move the center of mass, and the legs moving to locomote the robot, we were able to produce a walking motion. \cite{umer2016lbms}

With this initial design, a stable walking was generated in a heuristic manner. To generate the walking motion in a more methodical manner, further simplification was needed for ease in modeling and control design. As shown in Fig. \ref{walker}, two more design iterations were carried out. At first, the tray structure was simplified to a planar version where the ball was only allowed to roll along the length of the tray, while the tray could still do the pitching movement. In the second iteration, however, the pitch angle of the tray was restricted such that the ball then could move only in the sagittal direction. With this simplified design, it was much easier to model how the legs and the core behaved with respect to each other. For this structure, control strategies were designed successfully and it was shown that walking could be achieved.  \cite{umer2016control} In the next iteration, a three dimensional structure is being developed for which these control strategies will be investigated.

\subsection{Translating Intention-Based Motor Behaviors to Models in the Social Environment}

	When creating robotic systems for use in human environments, the role of movement arrives at the forefront of communication. Designers will need to consider the perception of any given system within the social context as well as acknowledge adaptability to, and participation in, dynamic environments. To approach a method of choreographic application and movement analysis of and for a robotic system, dance and body knowledge provides a necessary top-down perspective to uncover learned motions and the intricacy of motor control observable in complex life forms.
	A benefit of researching movement of and for a robotic system is exposed in the personalization and adaptation of social technologies. When people communicate with one another, mimicry of movement and emotion can be observed; this mimicry can otherwise be termed the chameleon effect \cite{lakin2003chameleon,chartrand1999chameleon}. Movement mimicry and emotional mimicry are observed through differing methods with movement mimicry retaining objectivity and the latter displaying subjective affect \cite{hess2013emotional}. It is possible that LBMS is able to characterize both movement and emotional mimicry with its attention to Body, Effort, Shape, and Space. In effort, Affinities, alongside Efforts, have been incorporated into the translation of breathing movements into/onto an animated virtual model.
	
	Considering bioinspiration and biomimicry as existing design methods for system development, we began with the movement of breath. Beginning with breath is a commonplace origin for dance movement studies as it is one of the most essential movement patterns within the body that is both unconscious but can be modified through conscious control. When someone feels calm, breathing may manifest as slow and released. When emotions are high, breath can become quicker and labored. When designing a model for use in human spaces, the movement of the model must be considered with mimicry in mind in order to reflect appropriate reactions to stimuli or to encourage necessary reactions in the people nearby.
	
\begin{figure}[h!]
\centering
\vspace{-.1in}
\includegraphics[width=.98\columnwidth]{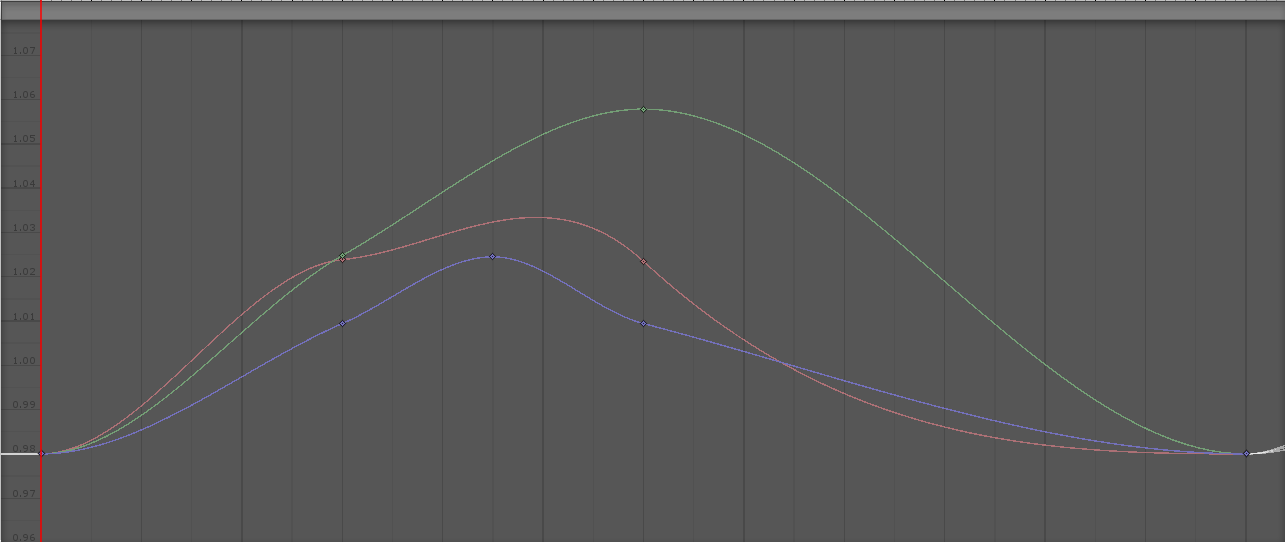}
\caption{Baseline animation curves graph. (a) $x$-curve is displayed in pink. (b) $y$-curve is displayed in green. (c) $z$-curve is displayed in purple.  These curves were crafted by a dancer to match her internal experience of breath inside different emotive states.}
\label{reika}
\end{figure}

	When developing breath of/for the virtual model, the researcher, and dancer, began by examining her own breathing patterns with somatic training in mind. Self-analysis of a baseline breathing pattern was conducted on the floor in a constructive rest position: a supine orientation was taken with the plantar surfaces of the feet resting on the ground, the knees falling in toward each other, and the leg muscles relaxed; the arms were placed palms-down onto the ventral surface of the body. Imagining first breathing into the lower cavity of the abdomen, toward the dorsal surface, into the thoracic cavity, and up near the clavicles. {Figure \ref{reika} represents the product of the student’s self-analysis of baseline breathing patterns. The breathing pattern was translated onto a virtual model that was created in the Unity video game engine. 
	
When analyzing the curves graph represented in Figure \ref{reika},} it is possible to see the rising and falling of the curves in the $x$, $y$, and $z$ directions; $x$ falls along the coronal plane, $y$ along the sagittal plane, and $z$ is perpendicular to the transverse plane. Due to the nature of individual differences and tremendous amounts of possibilities for any movement, variance would occur when mapping breathing patterns in any given person as well as for each consecutive breath. This particular person shows the greatest amount of perceived variance along the $y$-axis. In addition, the $x$-value increases sooner than the other values and the $z$-value shows the greatest limitations in range. Further analysis of other individuals may allow for general patterns to arise. After the creation of the base-breath animation, further studies on the embodiment of Effort drives were explored and displayed into/onto the model as a method of transferring felt intentions of subsequent breathing patterns.

A model of this kind can also be used to train self-awareness in individuals looking to harness the benefits of attending to movement in system design. The model described in this section was used by a group of individuals; a description of the task provided and results are as follows: The individuals were asked to sit in a chair, with their feet planted firmly onto the ground, and to attend to their posture by feeling their sit bones (ischial tuberosities) making contact with the chair. They were then instructed to attend to their own breathing. After attending to their breathing for some time, they were asked to breathe in the $x$, $y$, and $z$ directions, of the model’s orientation, which correspond to their body's Vertical, Sagittal, and Horizontal Dimensions, one at a time. After such introspection, the lab members were then asked to watch and mimic the spatial intentions seen in looped animation of the sphere model performing the base-breath breathing pattern determined by the previously-provided curves graph. After allowing several minutes to feel the subtle manifestations of breath in movement, each person contributed to a discussion on the observations they may have had throughout the exercise. Many members stated or agreed with feeling a sense of being calm, feeling more relaxed, and being more aware of or sensing a magnification of their personal movement patterns.

\subsection{Design of an In-Home Robot Companion}\label{ishaan_sec}
How expressive does a person, or a system have to be, in order to become a character in a story? How many modalities does it take to embody a character? Inherently, everything is expressive in some sense. Just how expressive something is depends on the number of modalities available to describe it. In robotic systems that mimic human expression, social intelligence and character traits can be established through the use of facial expressions, as is seen in anthropomorphic robots such as Kismet \cite{breazeal2003emotion}. Non-anthropomorphic systems, however, cannot rely on facial expression. These systems are more limited in their modalities. To tackle this problem, researchers at the RAD Lab proposed a design methodology that can be used to abstract archetypal characters across the depth of the design and interface elements of a non-anthropomorphic robotic system. This methodology uses the idea of perceived social intelligence in robot systems, whereby users associate a certain level of intelligence with the characteristics exhibited by the system in question. In \cite{bates1994role}, the authors write, ``There is a notion in the Arts of {`}believable character.{'} It does not mean an honest or reliable character, but one that provides the illusion of life, and thus permits the audience{'}s suspension of disbelief .... Traditional character animators are among those artists who have sought to create believable characters ....{"}. Thus to design effective robot characters, it is important that the characters themselves are believable. To this effect, this project made use of character archetypes. Archetypes are common blueprints for character traits that can be found in stories across cultures, and are thus easily identifiable. For instance, in A. A. Milne's Winnie the Pooh \cite{milne1977world}, each character represents a vivid archetype. Pooh is an archetypal easy-going character, Tigger is sanguine, happy and chirpy whereas Eeyore is melancholic and depressing. The characters Tigger and Eeyore are used to explore this character abstraction. To abstract these characters onto a robotic system with limited modalities, the researchers used the Kansei Engineering Flow iterative design approach \cite{nagamachi1995kansei}.

Kansei Engineering was first introduced as a design methodology in Japan in the 1980s. The Japanese word ``kansei'' refers to a ``consumers psychological feeling towards a new product'' \cite{nagamachi1995kansei}, and Kansei Engineering presents an iterative approach towards the design of consumer-focused products. This methodology involves a category classification through a top down approach. First, a zero level concept is chosen that outlines an intended high-level outcome. This zero level concept is broken down into clear subconcepts. Each subconcept is then further broken down into successive subconcepts that takes us from the intended high-level outcome, to a more specific description of the physical specifications of our product. By applying the Kansei Engineering design methodology to characters Tigger and Eeyore, the team arrived at a design for two aerial robots that represent each archetype. The proposed design is an aerial robot with a colored Styrofoam encasing, along with two LCD displays that showcase animated eyes. This allows the researchers to use color, expressive eyes and movement profiles as modes of expressing character traits. The movement profiles of the system are guided by the taxonomy of LBMS to study and observe the movement of each archetype. The researchers used the descriptive words, ``bouncing'' for Tigger and ``dragging'' for Eeyore, where a light motion quality for Tigger's bouncing and a heavy quality of Eeyore's dragging are correlated to Light Weight Effort and Strong Weight Effort. In humans these qualities are expressed through muscle tonus, the relationship of the pelvis to the ground, and other factors that cannot be recreated on our platform. For this system, the affinity with the Vertical dimension is used to put the Tigger profile spending more time higher up in space while the Eeyore profile spends more time lower in altitude. Thus, bouncing is expressed in a movement profile described by the sinusoid $|\sin(t)|$ with a greater amplitude and faster cadence. Dragging is expressed in a motion profile described by the sinusoid $1-|\sin(t)|$. A third, control aerial robot was designed to act as the baseline. This robot has no animated eyes, and is gray in color. Additionally, it's movement profile does not include a ``bouncing'' or ``dragging'' motion, and it does not hover with a sinusoidal cadence.

The effectiveness of the proposed design methodology was tested in a user study in Virtual Reality, using the HTC Vive system. The robot characters, Tigger and Eeyore were abstracted onto an aerial robotic platform that is designed in a parametric 3D software (Autodesk Fusion 360) using results from the Kansei Engineering Methodology. The environment is designed in the Unity 3D game development platform to look like a household living room, with furniture, shelves and a television. The use of Virtual Reality allowed the researchers to implement aspirational motion profiles that can be tested before applying them to hardware. Additionally, it serves as a safe testing environment, as aerial robots can be dangerous, with high-speed rotating blades.

A group of 45 participants participated in the study, and were divided into two groups labeled primed and unprimed. The primed group was shown two 30 second videos of the characters Tigger and Eeyore from the Disney cartoon adaptation of Winnie the Pooh. These videos showcase the character traits of each character that fit their respective archetypes. Following this, both sets of participants were put in the Virtual Reality environment. The participants were then asked to play a game of tic-tac-toe with a robot opponent. By being engaged in a game, the participants are cognitively active while they observe characteristics of the opponent. From our results, the team observed that users are prompted to attach expressive intent to the robot{'}s movements and decision making based on the character backstory, thereby increasing the robot's likability and perceived social intelligence, and subsequently the likelihood of adoption. The participants rated the robot Tigger to have a more ``bouncing'' movement, and the robots Eeyore and control to have a more ``dragging'' movement. Additionally, the robot Tigger was seen as more happy, optimistic and relaxed, whereas the robot Eeyore was seen as sad, pessimistic and anxious. These metrics were significantly bolstered in the primed case for the robot Tigger. Thus, the researchers were successfully able to identify the design characteristics that can be used to imply different character traits on a common robot platform architecture with limited expressive modalities.

\begin{figure}[h!]
\centering
\vspace{-.1in}
\includegraphics[width=.98\columnwidth]{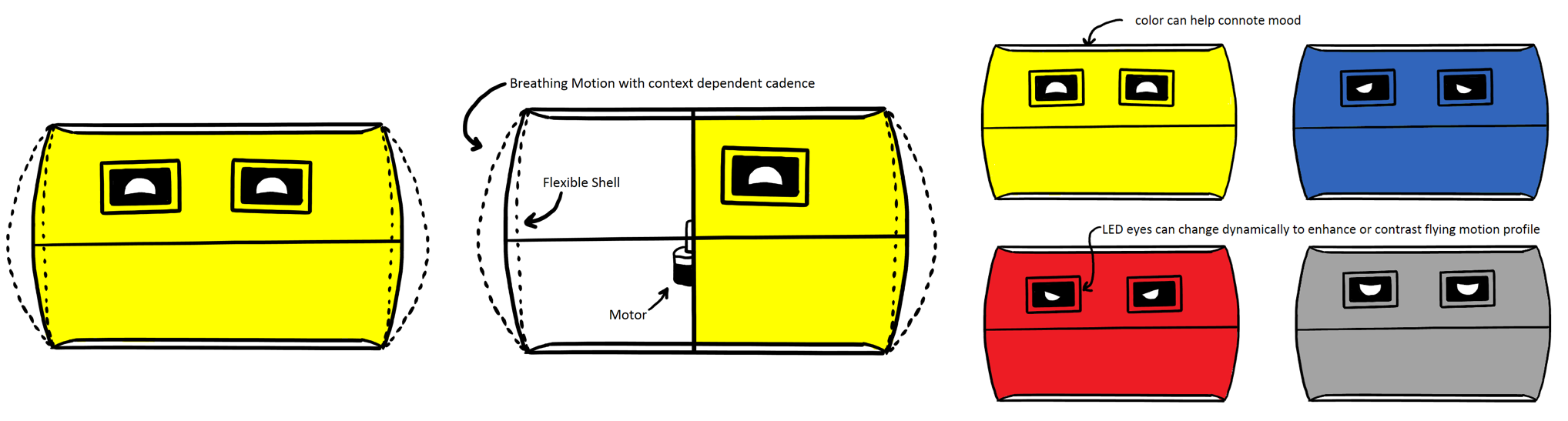}
\caption{Design for affective, expressive robot UAV for in home contexts.  Creating additional, dynamically unconstrained degrees of freedom will allow the robot to communicate internal system state in a visible fashion for in home users.}
\label{ishaan}
\end{figure}

This methodology, inspired by embodied movement exercises allowed us to look for ways to abstract character traits onto robotic systems with limited expressive modalities. Future work will explore the creation of less commonly used character tropes and how priming can impart these more subtle character creations effectively, with the hope of extending this work to design and build real aerial robotic systems as mocked up in Figure \ref{ishaan}. This will be accomplished alongside artists who will help develop characters for contexts of interest. Such studies can help create mechanized systems that are better accepted inside human-facing contexts and ensure that operational conventions are well communicated to users.

\subsection{Impact on Subjective Experience (Through an Interactive Installation and Performance)}

 
 The successful creation of meaningful and impactful artificial characters is dependent on the context provided to the observer. In specific, the subjective interpretation of the movement of these characters is thought to be heavily swayed by context. 
In this project we worked to assemble a malleable environment that a choreographer could use to dramatically change the expression of through choreography, lighting elements, musical accompaniment, costume, and other theatrical elements.
The idea of an interactive installation came forth as the goal of the collaboration, a means to manipulate the context given to the observer and test the effects it would have on the interpretation of the movement of select artificial agents. The experience would be further augmented by the inclusion of a performance involving a dancer, the Baxter, and the NAO. 

The work in progress resulted in a few key themes to continue exploring.  1) ``The Hidden Human Network'': Many technologies are powered by humans for the benefit of each other, but often this network is occluded, leaving a machine seeming quite intelligent, e.g., IBM’s Watson, which is powered by the webpage postings of users all over the Internet.  2) Are humans becoming more robot-like? This question was originally posed in the reverse, but upon further inspection, it is easy to argue that the rich adaptability of humans is heavily exploited in emerging technologies (more than any particularly successful imitation of biology).  With these changes, are humans finding social structures like family or friendships in embodied and personal technology experiences?  3) ``Time to Compile'': How long does it take to find resolution with or understanding of different technologies?  How long before we iterate on the first design and find a second?  Who gets to investigate the inner-workings of these machines before?  When have we assimilated a new technology permanently? How will we change?


\begin{figure}[h!]
\centering
\vspace{-.1in}
\includegraphics[width=.78\columnwidth]{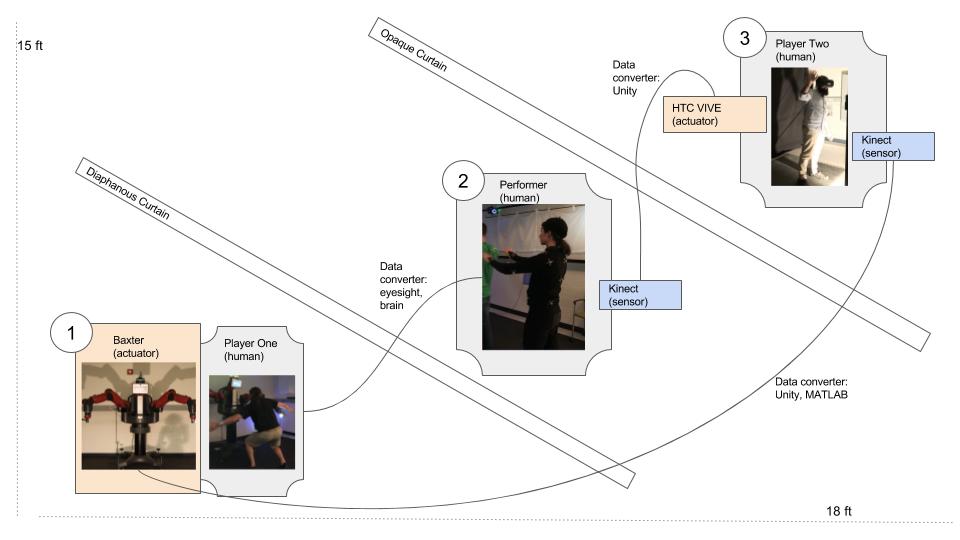}
\caption{Design of interactive performance loop with images of participants in June 30, 2017 public showing.  Future iterations of this loop will include more novel presentations of robots, through physical adornment, priming backstories, and motion design.}
\label{catie}
\end{figure}

The installation, shown in Figure \ref{catie}, took the form of an interactive loop composed of four unique interactions with four different artificial agents. The first interaction would be with the Baxter, wherein the participant would be asked to follow the movements of a flashlight held in Baxter’s hand with their own. The participant would effectively be mirroring the hand movements executed by Baxter. Standing directly behind the participant, but obscured from view by a translucent curtain, would be the performer, wearing a motion capture suit. She would attempt to mirror the movements of the participant precisely, with this motion data being relayed to a second participant wearing an HTC Vive. This participant would see the performer's movements in the motion capture suit represented one-to-one by a virtual avatar in the VR environment. They would be instructed to mimic these movements exactly, to be captured by a Kinect and relayed to the Baxter as motion commands. The movements of the second participant, therefore, would dictate the movements made by Baxter, which dictated the movements made by the first participant, and so on. In this way, the feedback loop was closed and eventually, it could be seen that all agents in the loop, both artificial and human, would be performing the exact same sequence of movements with only a slight delay between them. 

The interactive experience is precluded by a choreography between the performer and the robots; this designs a priming experience that will set the tone for how participants interact with the machines in the loop. At the end of the entire experience, users are asked to draw how they believed the different sections of the installation to be related. 
Interpretation
 is malleable based on context and relationship.  The interactive  installation creates a context where the central human character primes the audience about the artificial agents including a Baxter and a NAO robot.  For most audience members, this is their
 first experience viewing these robots and seeing them interact with another human (the performer).  As the relationship between the human character and the robots progresses throughout the plot, one goal is for the audience to recognize that the human controls
 the robots, therefore rendering them as subservient as a hammer or wheel. 
Through this experience, we hoped to seed the formation of certain questions in the mind of the participant. Do you feel empathy for the robots? Are robots becoming more human, or is it the other way around?

  



%
%

\section{Discussion: Establishing Principles for Interdisciplinary Study Between Robotics and Dance}\label{discussion}
This paper has provided background on tools and techniques from choreography and somatics and how those have been used in the development of more expressive robotic systems.  We've posited that a missing piece from much of the prior work in robotics is meaningful interdisciplinary collaboration, which cannot happen without embodied practice.  To that end, we've presented methods developed in the RAD Lab with our many collaborators to provide embodied experiences for an interdisciplinary group working in robotics.  To show that these approaches can be effective and spur innovation, we've shared overviews of projects that have benefited within our group.  Finally, as a result of these collaborations, in this section we offer some over-arching conclusions to help guide others in similar practices.  

\subsection{Framing of Activity, Values, and Norms}

        
Embodied practice in an engineering context must address the challenge of varying experience levels on both sides.  Dancers are accustomed to specific taxonomies like LBMS or the vocabulary of ballet.  They create and recreate movement phrases nearly instantaneously on their path to a final finished product.  On the other hand, engineers utilize language like ``nondeterministic'', a phrase that a dancer might loosely (and unsatisfyingly) equate to ``improvisation''.  It may take hours for one version of a coded system to work once.   Interdisciplinary activity creation thus requires a shared, dynamic  vocabulary and respect for process on the part of the instructor and the students.  This requires classifying certain choreographic devices like repetition or retrograde as ``techniques within technologies''.  The act of extrusion in a CAD software is a technique inside of a technology used by engineers in the same way that body positions in the style of Vagonova ballet is a technique within a technology used by choreographers.  The notion of ``experience'' extends beyond language, whereas different students have varying backgrounds in activities like singing, jumping, or making physical contact.  This poses a challenge of creating exercises where all students can participate fully and where exploration is not contingent upon experience level.  This challenge spread to the collaborations in research where a lack of experience with a tool like MATLAB will inhibit a choreographer's ability to direct robot motion.  Navigating this ambiguity requires a safe space for honest communication and respect for the respective ``compile times'' of distinct processes.

Moreover, it's a divide that can be reinforced by the financial side of collaboration.  Economic associations, salaries characteristic in each field, the ``hardness'' of each as undergraduate majors, distribution of public funds (at every level of education as well as in research), and other gendered and power-wielding norms common in Western society put these endeavors at an immediate uneven footing -- with engineers having the upper hand in nearly every feature.  It is easy for participants to view dance curriculum as lesser than mathematics, physics, and mechanics courses.  It is easy for many academic audiences to judge embodied exercises to be lacking intellectual rigor or relevance.  But, when those preconceived notions can be set aside, the discoveries for both groups will be invaluable.  Nontechnical, or qualitative, movement and thinking does not equate to easy or simple; moreover, as this paper has illuminated, there is quite a bit of ``technique'' involved.  In addition to navigating the interpersonal dynamic that this difference can create, this means that financial support and intellectual attribution must be provided for all participants as further discussed in the next section.

\subsection{Logistics of Working with Dancers, Choreographers, and CMAs}
 In these examples we see the desire for subjective influence on human counterparts, which requires objective observation of movement patterns in context.  This is an area where not only do we lack the quantitative methods to produce consistent, meaningful analysis of movement, but also the act of qualitative observation can be tricky.  Indeed, this is what CMAs (and CLMAs) study and practice for years to develop proficiency.  Viewing movement is complex and takes place inside an inherently biased human form.  Part of the training provided in LBMS programs is in identifying one's own biases and movement patterns in order to see more clearly.  This necessarily means that observation by a trained expert needs to be marked and tracked through academic work -- both to attribute expertise and to acknowledge that a different scholar might come to different conclusions.  Recent publications outside our group reinforce the value of collaboration with movement professionals and highlight a troubling practice of not honoring expert observation with paper authorship \cite{salaris2017robot}.
 
Leaving ``nontechnical'' collaborators off of co-author lists in robotics also reinforces the lack of funding for this type of work as practitioners do not build an academic track record to bolster future efforts.  Indeed in every example on this paper, the engineering side of the collaboration controls the majority of the funding for the joint work.  This imbalance feeds into the cultural issues discussed in the prior section.  In our group, we work to communicate consulting and teaching hourly rates for collaborators in a transparent, up-front manner.  Currently, the rates we use in our work are \$150/hour for teaching (and preparing for) group classes and \$70/hour for consulting.  It requires unreasonable patience on the part of our collaborators outside of university engineering departments who might not be reimbursed for their work for months due to the structure of university reimbursements.  Instruments that would allow these collaborators to be paid in a timely fashion would help at the university level.  Empowerment of these parties to pursue and direct their own research need to be addressed by federal agencies and institutions where collaborative efforts can be acknowledged through independent awards to working dance artists and somatics professionals.


\subsection{Irreplaceable Body-Based Research in Robotics}
As we have motivated here and as engineer after engineer who has participated in the describe activities has attested, much of this knowledge cannot be learned from a textbook.  On the one hand, the LBMS community does not regularly publish about their work (in addition to not always being offered co-authorship on publications on which they work as discussed above); this is something that has to be addressed within that community -- and that funding to that community as motivated in the previous section -- could help diffuse.  On the other hand, this work is of the body, not the mind, and is necessarily passed on through embodied practice.  As a result, many of publications in robotics that leverage the system lack in depth knowledge of it, often referring to it as emotion or body language analysis. Few researchers in robotics are CMAs or CLMAs.  Yet, roboticists (including reviewers at publishing venues) are quick to challenge our nuanced understanding of the system, saying ``The authors' understanding of Laban Movement Analysis is by far insufficient. Some statements are just not true. For example .... Flow [Effort] does not depend on degrees of freedom at all.''  This kind of mechanistic assumption by a roboticist, who likely is not a CMA or CLMA, of quantitatively what Flow Effort \textit{is} -- rather than how this phenomenon might be modeled or approximated -- is indefensible and impedes the ability for the engineering community to learn from the expertise of outside fields.  To address this, more roboticists need embodied training in choreography, somatics, and LBMS; through working in these methods, we see first hand how complex human movement and the patterns it expresses are.  As described in Section \ref{umer_sec}, this can lead to novel robot designs that show promise in improving the functional (and thus expressive) capabilities of robotic platforms.  Just as biomedical researchers collaborate with medical practitioners and visit hospitals to understand the activities and stakeholders within, roboticists building expressive systems for human interaction need to spend time in a dance studio working with and learning from choreographers and dance professionals.  


%

\subsection{It Has to Be ``Ad-Hoc''}
The idea that artificial movement generation not based on a physical model or optimization is ``ad-hoc'' and thus invalid is at odds with the perspectives described here.  For example, in work such as  \cite{knight2014expressive,etemad2016expert} where specific models are given for movement that is ``happy'' and ``sad'' our follow on work \cite{madi2017ICSR} has indicated these models to be limited to a particular context.  Moreover, spending time in a dance studio with a choreographer will illuminate how malleable motion interpretation can be: the same dance in a different costume will generate a different mood or feeling in viewers.  Traditional robotics venues have balked at creative tools and methodologies for artists as being too ``ad-hoc'' with reviewers offering constructive feedback like ``The key postures are also manually provided for different styles -- this does not sound very scientific.  I think it is necessary to automate such process to claim a contribution on stylizing the movements'' and ``if a different designer had done the transformation, very different characters might have emerged. The design process is highly subjective and does not generalise''.  However, we think these are \textit{strengths} of our methods.  Again, we point out that the many myriad of ways human motion may be translated to a lower degree of freedom system, requiring some ad-hoc choice in the process;  we posit that premature automation and restrictive optimization ends up limiting the exhibited behavior of robotic systems.  On the other hand, think of the many myriad ways motion as simple as a Roomba has been interpreted in the inherently varied contexts it inhabits in human homes.  Offering this choice to a human (or choreographer) is the only way to empower collaborators in dance.  Moreover, based on all the ways that context (environment, lighting, backstory, culture, slang, to name a few) informs movement reception, we posit there is no way to fully automate this process.  Work in computer science has been offered similar latitude and produced useful tools \cite{hudak}.

\subsection{Objective, Qualitative Movement Observation (to Support Subjective Conclusions)}
\begin{figure}[h!]
\centering
\vspace{-.1in}
\includegraphics[width=.98\columnwidth]{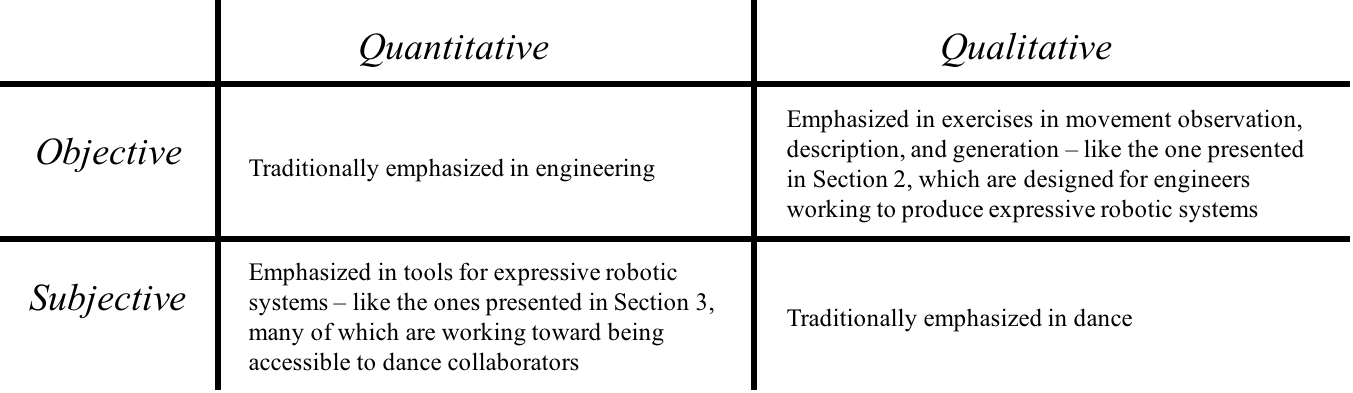}
\caption{Qualitative and quantitative methods are needed to explain and design objective and subjective phenomena.  This diagram frames how the methods and results presented here aid in this need, bringing balance to the upper right and lower left quadrant of the diagram.}
\label{quad}
\end{figure}

In pursuit of automation that is sensitive to movement expression, then, we motivate the need for a pipeline of activity which starts with observation of movement.  Only through noticing the objective mechanisms at play with qualitative methods (e.g., written description) can we begin to quantify and automate aspects of expression through movement.  The diagram in Figure \ref{quad} highlights this take away, motivating the need for two areas of future work: qualitative methods for the description of objective movement phenomena and quantitative methods for the analysis of subjective phenomena.  The former is a way of characterizing all of the methods presented in Section \ref{methods}, while the latter describes the results presented in Section \ref{results}.  For example, every conversation spurred on by the embodied practices employed here is a chance for engineers and dancers to describe the experience of moving in their own bodies and viewing movement in others.  On the other hand, every output of expression through movement that interfaces with some technological element or another requires quantification.  

These lenses also reveal how important objective description is to seeing mechanisms in movement that can be translated to automated systems.  It is tempting for engineers to watch an animate character that may serve inspiration for a robot companion and say: ``I'm going to replicate this happy movement in my robot design''.  As explained in Section \ref{ishaan_sec}, this lacks objective explication of movement phenomena that can be translated to an engineered system.  Instead engineers need to break down the behaviors that are contributing to that character's ``happy'' affect within its given context (Section \ref{madi_sec}).  In this light, an engineer can then see: ``Ah, the frequency of motion in the vertical dimension is modulated when the character's internal state varies, and an increased frequency of the period of the motion correlates with a happier seeming character''.  This can be translated into quantitative design elements:  period of oscillations.  Moreover, it motivates an important separation between ``style'' (how a platform is moving) and ``affect'' (how a human responds to motion).  Expressive robotic systems will have wide capacity for a range of motion styles that, when aspects of context and characteristics of a human counterpart are accounted for, can communicate a desired affect to a human counterpart.

\section{Conclusions: Toward Widespread Collaboration Between Dancers and Roboticists}\label{conclusion}
\begin{center}
\textit{...the truest creativity of the digital age came from those who connected the arts and sciences.}
\end{center}
\begin{flushright}
-- The Innovators \cite{isaacson2014innovators}
\end{flushright}

There is inherent excitement around collaborations between engineering and the arts -- as evidenced by the quote above.  As cheaply and easily as that excitement is generated, it can impede more meaningful discourse between two very different disciplines.  For example, our lab is often inaccurately (or jokingly) characterized as creating ``dancing robots'' -- of value only for entertainment purposes.  This occurs because of the chasm that typically divides these fields.  Not only are the methods in each distinct but so too are the culture, values, norms, outputs, and resources of each field.  However, robotics is at a critical point where researchers are working to bring complex mechanical machines outside of controlled factory environments, where they are walled-away from human counterparts.  This demands new understanding of how humans view and create movement that experts like dance artists, choreographers, and somatic practitioners already understand.  This article has outlined a variety of methods that can help that process and reviewed examples of impact in the robotics field that these experts are already having.  Broadly speaking, these outcomes work toward the production of expressive robotic systems:  systems that have a greater variety of choice in the generated motion profiles.  This makes it easier for humans to design movement.  It makes it easier for humans to interpret movement.
Is such a system more life-like? Yes, because it’s more expressive. Is it dancing? No, that’s the humans operating it, designing it, selecting moving patterns over time, and extending themselves through technology. 
Thus, as a more refined paintbrush or a camera with better dynamic range, with improved, more expressive robotic technology, the machine \textit{extends the artist} into the 21st century. 

\vspace{6pt} 


\acknowledgments{The work in this paper was sponsored by NSF grant numbers 1528036 and 1701295, DARPA grant number D16AP00001, a grant from the Jefferson Trust, a grant from the UVA Data Science Institute, and start up funds from UVA and UIUC.  We thank all of our collaborators, including those who could not contribute directly to this paper, who have helped us along the way.}

\authorcontributions{Each co-author contributed directly to at least one section of the paper.  Authors with a $^\dagger$ designation wrote one or more sections themselves; authors with a $^\ddagger$ designation contributed significant chunks of text, heavy edits, and important main ideas.   The lead author compiled, organized, and wrote the several sections of the paper.  Please see the list in Section \ref{intro} for further detail about author roles.}

\conflictsofinterest{The authors declare no conflict of interest.  The founding sponsors had no role in the design of the study; in the collection, analyses, or interpretation of data; in the writing of the manuscript, and in the decision to publish the results.} 


\appendixtitles{no} 
\externalbibliography{yes}
\bibliography{papers}


\end{document}